\pdfoutput=1

\documentclass[11pt]{article}

\usepackage{ACL2023}

\usepackage{times}
\usepackage{latexsym}
\usepackage{amsmath, amssymb}
\usepackage{graphicx}

\usepackage{array}
\usepackage{booktabs, dcolumn, caption, bm}
\usepackage{multirow, multicol}
\usepackage{subcaption}
\usepackage{makecell}
\usepackage{xcolor,colortbl}

\usepackage{algorithm}
\usepackage{algpseudocode}
\usepackage{amsthm}
\usepackage[outline]{contour}

\usepackage{float}
\usepackage{tabularx}
\newcolumntype{d}[1]{D{.}{.}{#1}} 
\newcommand{\bigcell}[2]{\begin{tabular}{@{}#1@{}}#2\end{tabular}}

\usepackage[T1]{fontenc}

\usepackage[utf8]{inputenc}

\usepackage{microtype}

\usepackage{inconsolata}

\usepackage{multirow}

\title{\textsc{Tapir}: Learning Adaptive Revision for Incremental Natural \\ Language Understanding with a Two-Pass Model}

\author{Patrick Kahardipraja$^{\mathbf{1}}$ \hspace{10mm} Brielen Madureira$^{\mathbf{1}}$ \hspace{10mm}  David Schlangen$^{\mathbf{1, 2}}$ \\
$^{\mathbf{1}}$Computational Linguistics, Department of Linguistics \\ University of Potsdam, Germany \\
$^{\mathbf{2}}$German Research Center for Artificial Intelligence (DFKI), Berlin, Germany \\
  \texttt{\{kahardipraja,madureiralasota,david.schlangen\}@uni-potsdam.de}}

\begin{document}
\maketitle
\begin{abstract}

Language is by its very nature incremental in how it is produced and processed. This property can be exploited by NLP systems to produce fast responses, which has been shown to be beneficial for real-time interactive applications.
Recent neural network-based approaches for incremental processing mainly use RNNs or Transformers. RNNs are fast but monotonic (cannot correct earlier output, which can be necessary in incremental processing).
Transformers, on the other hand, consume whole sequences, and hence are by nature non-incremental.
A \emph{restart-incremental} interface that repeatedly passes longer input prefixes can be used to obtain partial outputs, while providing the ability to revise. However, this method becomes costly as the sentence grows longer. 
In this work, we propose the Two-pass model for AdaPtIve Revision (\textsc{Tapir}) and introduce a method to obtain an incremental supervision signal for learning an adaptive revision policy. Experimental results on sequence labelling show that our model has better incremental performance and faster inference speed compared to restart-incremental Transformers, while showing little degradation on full sequences.\footnote{Our implementation is publicly available at \url{https://github.com/pkhdipraja/tapir}.}
\end{abstract}

\section{Introduction}
\label{sec:intro}

Incrementality is an inseparable aspect of language use. Human speakers can produce utterances based on an incomplete message formed in their minds while simultaneously continuing to refine its content for subsequent speech production \citep{kempen-hoenkamp-1982-incremental, incrementalproceduralgrammar}. They also comprehend language on (approximately) a word-by-word basis and do not need to wait until the utterance finishes to grasp its meaning \citep{anticipatory}.

\begin{figure}[t]
	\centering
    \includegraphics[width=\columnwidth,trim={0cm 16.3cm 11.5cm 0cm},clip]{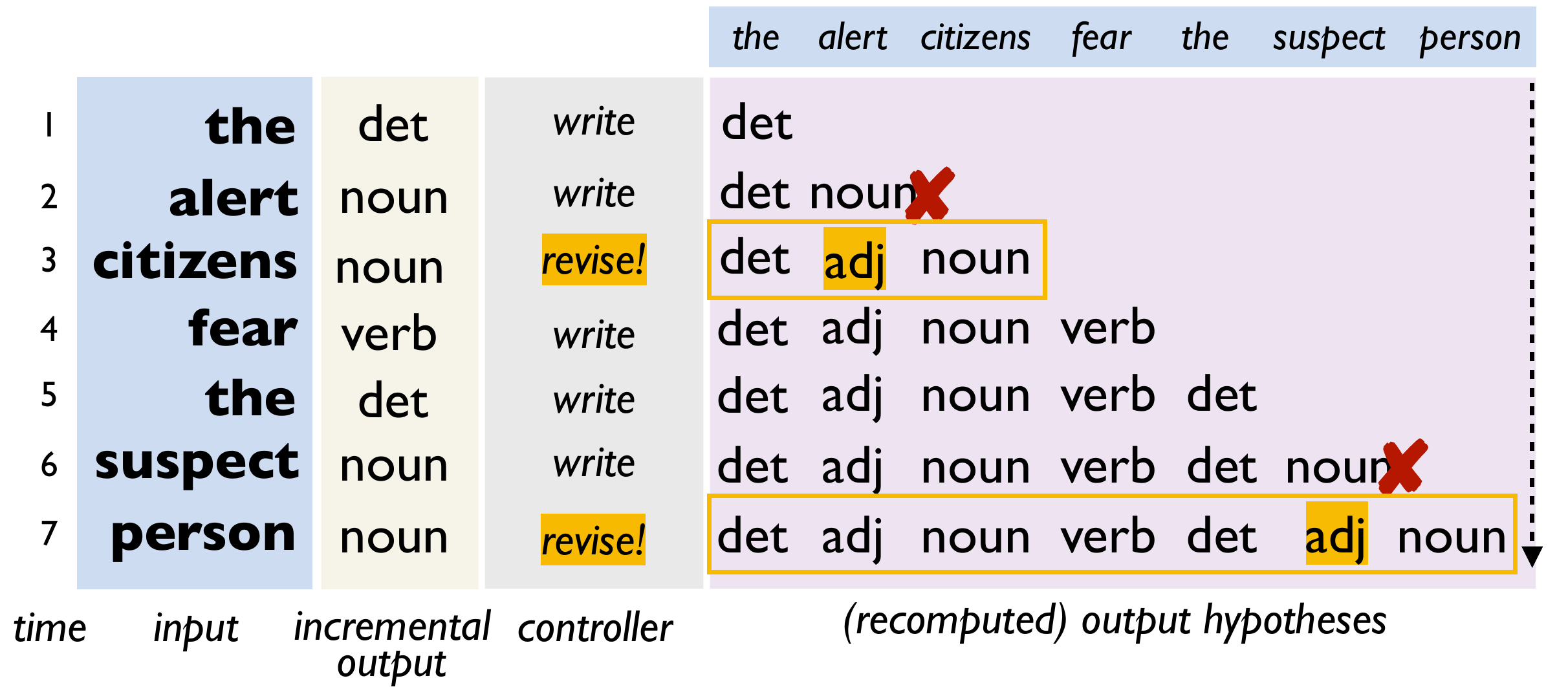}
	\caption{Illustrative example of how a monotonic incremental POS-tagger would not recover from wrong hypotheses. A policy for adaptive revision, here parameterised by a controller, can enable reanalyses to be performed when necessary (here at time steps 3 and 7).}
	\label{fig:inc-proc-example}
\end{figure}

As observed by \citet{madureira-schlangen-2020-incremental},
a natural option for neural network-based incremental processing would be RNNs \citep{Rumelhart1986LearningRB}, as they have essential properties required in incremental scenarios: They keep a recurrent state, are sensitive to the notion of order and are able to accept partial input and produce an output at each time step. Ideally, an incremental processor should also be able to revise its previous incorrect hypotheses based on new input prefixes \citep{schlangen-skantze-2009-general}. However, RNNs are unable to do so as their output is monotonic.

The Transformer architecture \citep{transformers} has been the \emph{de facto} standard for many NLP tasks since its inception. Nevertheless, it is not designed for incremental processing as the input sequences are assumed to be complete and processed as a whole. A \emph{restart-incremental} interface \citep{beuck-etal-2011-decision, schlangen2011general} can be applied to adapt Transformers for incremental processing \citep{madureira-schlangen-2020-incremental}, where available input prefixes are recomputed at each time step to produce partial outputs. Such an interface  also provides the capability to revise existing outputs through its non-monotonic nature. Although feasible, this method does not scale well for long sequences since the number of required forward passes grows with the sequence length.\footnote{Processing a sequence of $n$ tokens once turns into processing $n$ sequences with $\sum_{k=1}^nk$ tokens in total.}
The revision process is also not effective as it occurs at every time step, even when it is unnecessary.

Revision is crucial in incremental processing, as it is not always possible for a model to be correct at the first attempt, either because the linguistic input is provided in its inherent piecemeal fashion (as shown in Figure \ref{fig:inc-proc-example}) or because of mistakes due to poor approximation. One way to improve the output quality is the delay strategy \citep{beuck-etal-2011-decision, baumannetal2011}, where tokens within a lookahead window are used to disambiguate the currently processed input. However, it can neither fix past hypotheses nor capture long-range influences \emph{e.g.}\ in garden path sentences.

In this work, we propose the Two-pass model for AdaPtIve Revision (\textsc{Tapir}), which is capable of adaptive revision, while also being fast in incremental scenarios. This is achieved by using a revision policy to decide whether to \texttt{WRITE} (produce a new output) or \texttt{REVISE} (refine existing outputs based on new evidence), whose mechanism is described in \S\ref{sec:model}. Learning this policy requires a supervision signal which is usually not present in non-incremental datasets \citep{kohn-2018-incremental}. In \S\ref{sec:rev-signal}, we tackle this issue by introducing a method for obtaining action sequences using the Linear Transformer (LT) \citep{pmlr-v119-katharopoulos20a}. As silver labels, these action sequences allow us to view policy learning as a supervised problem.

Experiments on four NLU tasks in English, framed as sequence labelling\footnote{We do not run experiments on sequence classification, as revisions can trivially be performed by predicting one label at every time step.}, show that, compared to a restart-incremental Transformer encoder, our model is considerably faster for incremental inference with better incremental performance, while being comparable when processing full sequences. Our in-depth analysis inspects \textsc{Tapir}'s incremental behaviour, showing its effectiveness at avoiding ill-timed revisions on correct prefixes.

\section{Related Work}

There has been increasing interest to explore neural network-based incremental processing. \citet{lectrack} proposed a dialogue state tracker using LSTM \citep{lstm} to incrementally predict each component of the dialogue state. \citet{liu-etal-2019-referential} introduced an incremental anaphora resolution model composed of a memory unit for entity tracking and a recurrent unit as the memory controller. RNNs still fall short on non-incremental metrics due to their strict left-to-right processing. Some works have attempted to address this issue by adapting BiLSTMs or Transformers for incremental processing and applying it on sequence labelling and classification tasks \citep{madureira-schlangen-2020-incremental, kahardipraja-etal-2021-towards} and disfluency detection \citep{rohanian-hough-2021-best, chen-etal-2022-teaching}.

Our revision policy is closely related to the concept of policy in simultaneous translation, which decides whether to wait for another source token (\texttt{READ} action) or to emit a target token (\texttt{WRITE} action). Simultaneous translation policies can be categorised into fixed and adaptive. An example of a fixed policy is the wait-$k$ policy \citep{ma-etal-2019-stacl}, which waits for first $k$ source tokens before alternating between writing and reading a token. An adaptive policy on the other hand,
decides to read or write depending on the available context and can be learned by using reinforcement learning techniques \citep{grissom-ii-etal-2014-dont, gu-etal-2017-learning} or applying monotonic attention \citep{pmlr-v70-raffel17a, chiu*2018monotonic, arivazhagan-etal-2019-monotonic, Ma2020Monotonic}. 

The memory mechanism is a key component for revision policy learning as it stores representations which, for instance, can be used to ensure that the action is correct \citep{guo-etal-2022-turning}. It also absorbs asynchronies that may arise when each component in an incremental system has different processing speed \citep{levelt}. The memory can be internal as in RNNs, or external such as memory networks \citep{DBLP:journals/corr/WestonCB14, NIPS2015_8fb21ee7} and the Neural Turing Machine \citep{DBLP:journals/corr/GravesWD14}.

Revision in incremental systems has been previously explored. In simultaneous spoken language translation, \citet{niehues16_interspeech} proposed a scheme that allows re-translation when an ASR component recognises a new word. \citet{arivazhagan-etal-2020-translation} evaluated streaming translation against re-translation models that translate from scratch for each incoming token and found that re-translation yields a comparable result to streaming systems. \citet{zheng-etal-2020-opportunistic} proposed a decoding method for simultaneous translation that overgenerates target words at each step, which are subsequently revised. One way to achieve revision is by employing a two-pass strategy. \citet{deliberation} proposed a deliberation network for machine translation, composed of encoder-decoder architecture with an additional second-pass decoder to refine the generated target sentence. In dialogue domains, this strategy is also used to improve the contextual coherence and correctness of the response \citep{li-etal-2019-incremental} and to refine the output of retrieval-based dialogue systems \citep{ijcai2018p609, weston-etal-2018-retrieve}. Furthermore, the two-pass approach is commonly utilised in streaming ASR to improve the initial hypothesis \citep[\emph{inter alia}]{sainath2019twopass, hu-deliberation, DBLP:conf/icassp/WangHS22a}. 

The aforementioned works shared a common trait, as they used a fixed policy and performed revision either for each incoming input or when the input is already complete. Our approach differs in that our model learns an adaptive policy that results in more timely revisions. Contemporaneous to our work, \citet{kaushal-etal-2023-efficient} proposed a cascaded uni- and bidirectional architecture with an additional module to predict when to restart. The module is trained with a supervision signal obtained from comparing the model's prediction against the ground truth. Their approach is effective in reducing the required computational budget.

\section{Model}
\label{sec:model}

\begin{figure}[t]
	\centering
	\includegraphics[width=\columnwidth]{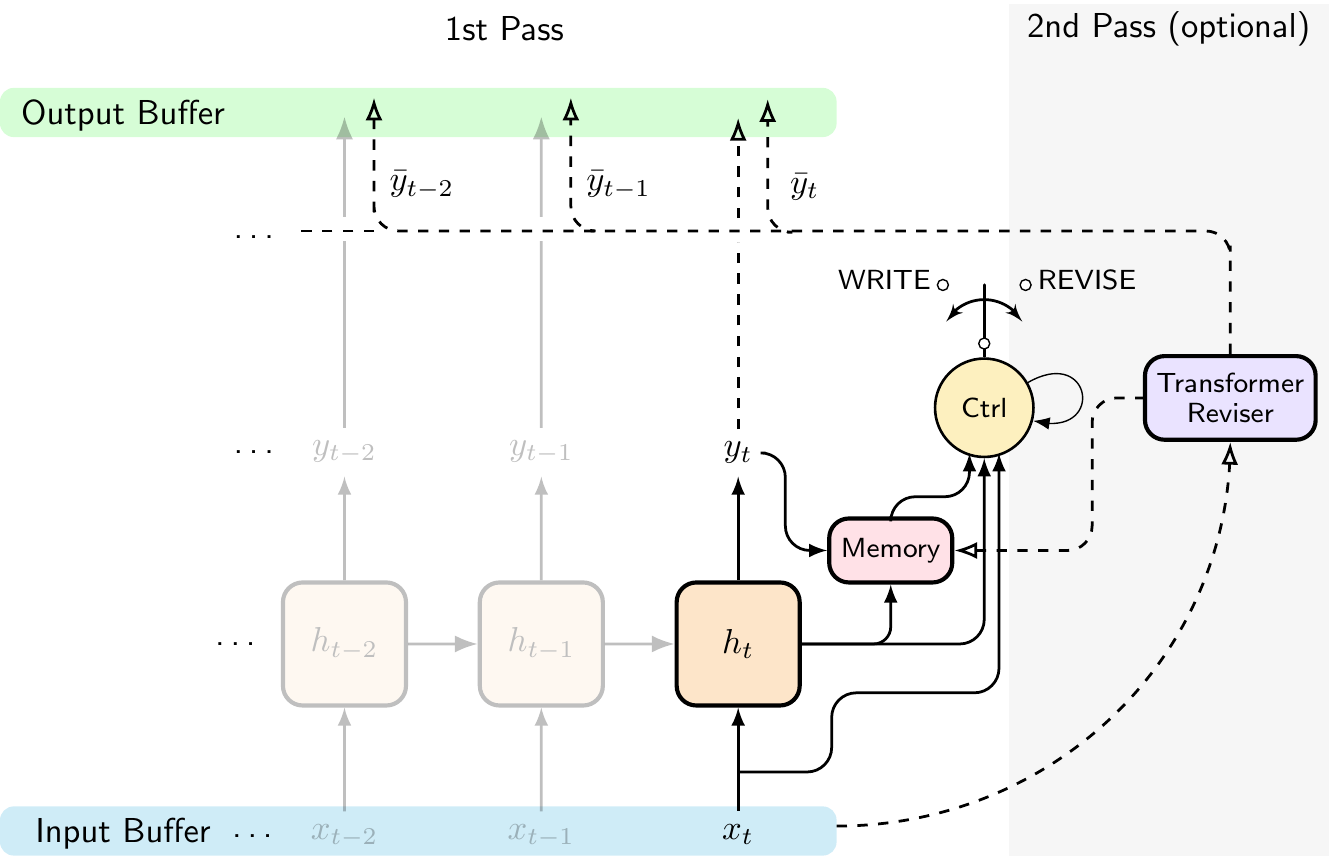}
	\caption{\textsc{Tapir} computes a candidate output using an RNN at each time step. Then the controller decides whether to \texttt{WRITE} by adding the new output to the output buffer or to take a \texttt{REVISE} action, which can edit the output buffer after observing the effect of the new input on past outputs with the help of the memory.}
	\label{fig:twopass}
\end{figure}

To address the weaknesses of RNN- and Transformer-only architectures for incremental processing (\S\ref{sec:intro}), we introduce a Two-pass model for AdaPtIve Revision named \textsc{Tapir}, which integrates advantages of both models and is based on the deliberation network \citep{deliberation}. Its architecture, depicted in Figure \ref{fig:twopass}, consists of four components as follows:

\begin{enumerate}
    \item \textbf{Incremental Processor}: a recurrent model that produces an output at each time step and serves as the first-pass model. In this work, we use a standard LSTM network.
    
    \item \textbf{Reviser}: a bidirectional model that can revise via recomputation operations (\S\ref{sec:policy}), also called the second-pass model. We opt for Transformer-based models following \citet{li20t_interspeech} as it allows parallel recomputation. The revision process corresponds to the \emph{forward reanalysis hypothesis} \citep{FRAZIER1982178}, where a sentence is processed from the beginning whenever the need for reanalysis is detected.
    
    \item \textbf{Memory}: the history of inputs and outputs are stored in the memory. Taking the inspiration from \citet{grave2017improving}, we use caches, as they are computationally cheap, offering a considerable speed-up in incremental settings.
    
    \item \textbf{Controller}: a neural network that parameterises the revision policy. We choose a recurrent controller following \citet{DBLP:journals/corr/GravesWD14}, as its internal memory complements the memory module and is also suitable for incremental scenarios. We use a modified LSTMN \citep{cheng-etal-2016-long} for this component.
    
\end{enumerate}

During incremental inference, \textsc{Tapir} computes a candidate output $y_t$ for the most recent input $x_t$ as the first pass. Then, based on $x_t$ and the memory, it decides whether to take a \texttt{WRITE} (add $y_t$ to an output buffer) or \texttt{REVISE} (perform a second pass to recompute all existing outputs) action. The action is defined by a revision policy $\pi_{\theta}$, which models the effect of new input on past outputs. At each time $t$, $\pi_{\theta}$ makes use of processed inputs $x_{\leq t}$ and past outputs $y_{< t}$ to select a suitable action $a_{t}$.\footnote{The output $y_{t}$ is excluded as it is not required to determine if a recomputation should occur in our model.} It is parameterised by the controller hidden state $k_{t}$ with a non-linear function $g$:
\begin{align}
    \pi_{\theta}(a_{t}|a_{< t}, x_{\leq t}, y_{< t}) \propto g_{\theta}(k_{t})
\end{align}

\subsection{Revision Policy}
\label{sec:policy}

In restart-incremental models, revisions can occur as a result of recomputations, which are costly since they happen at every time step, even when no revisions occur. \textsc{Tapir} revises by selectively deciding when to recompute, which enables it to revisit previous outputs at different points in time while reducing the number of recomputations.

\noindent \textbf{Memory Content}. The memory in \textsc{Tapir} contains information pertaining to processed inputs and their corresponding outputs, which is crucial for our approach. This is because it enables our model to perform relational learning between an incoming input and past outputs, using past inputs as an additional cue. Here, we use three caches $\Gamma$. $\Gamma^{h}$ stores the hidden state $h$ of the incremental processor, representing the current input prefix, $\Gamma^{z}$ stores the projected output vector $z$ which represents the output, and $\Gamma^{p}$ stores the input-output representation $\varphi$, which is computed from $h$ and $z$. The $i$-th slot of the caches contains $\gamma_{i}^{h}, \gamma_{i}^{z}, \gamma_{i}^{p}$, all of them computed at the same time step. The representations $z$ and $\varphi$ are computed as follows:
    \begin{align}
        z &= \tanh(W_{\tilde{y}}\tilde{y} + b_{z}) \\
        \varphi &= \tanh(W_{in}h + W_{out}z + b_{\varphi})
    \end{align}
where $\tilde{y}$ is the output logits from the incremental processor. $W_{\tilde{y}}, W_{in},$ and $W_{out}$ are parameters while $b_{z}$ and $b_{\varphi}$ are bias terms. The dimension of $z$ and $h$ is the same. We keep the cache size $N$ small, as we later perform soft attention over $\Gamma^{p}$. The attention computation for large cache sizes is costly and is not suitable for incremental scenarios. Due to this limitation, the oldest cache element is discarded when the cache is full and new partial input arrives.

\noindent \textbf{Modelling Actions}. To model possible changes in past outputs as an effect of a new input, we use an LSTMN controller due to its ability to induce relations among tokens. It computes the relation between $h_{t}$ and each cache element $\gamma_{i}^{p}$ via an attention mechanism:
    \begin{align}
        U &= W_{c}\gamma_{i}^{p} + W_{h}h_{t} + W_{\tilde{k}}\tilde{k}_{t-1} + b_{u} \\
        s_{i}^{t} &= \text{softmax}(v^{\top}\text{tanh}(U))
    \end{align}
which yields a probability distribution over $\Gamma^{p}$. $\tilde{k}_{t-1}$ is the previous summary vector of the controller hidden state. $W_{c}, W_{h}, W_{\tilde{k}},$ and $v$ are parameters and $b_{u}$ is a bias term. We can then compute adaptive summary vectors $\tilde{k}_{t}$ and $\tilde{c}_{t}$ as a weighted sum of the cache $\Gamma^{p}$ and the controller memory tape $C_{t-1}$: 
    \begin{align}
        \begin{bmatrix}
        \tilde{k}_{t} \\
        \tilde{c}_{t}
        \end{bmatrix}
         = \sum_{i=1}^{N}s_{i}^{t} \cdot 
        \begin{bmatrix}
        \gamma_{i}^{p} \\
        c_{i + \max{(0, t-N-1)}}
        \end{bmatrix}
    \end{align}
where $c_{i + \max{(0, t-N-1)}}$ is the controller memory cell for the corresponding cache element $\gamma_{i}^{p}$. The attention can be partially viewed as local \citep{luong-etal-2015-effective}, since older cache elements are incorporated through $\tilde{k}_{t-1}$. These summary vectors are used to compute the recurrent update as follows:
    \begin{align}
        \begin{bmatrix}
        i_{t} \\
        f_{t} \\
        o_{t} \\
        \hat{c}_{t}
        \end{bmatrix}
        &= 
        \begin{bmatrix}
        \sigma \\
        \sigma \\ 
        \sigma \\
        \text{tanh}
        \end{bmatrix}
        W \cdot [\tilde{k}_{t}, x_{t}]
        \\
        c_{t} &= f_{t} \odot \tilde{c}_{t} + i_{t} \odot \hat{c}_{t} \\
        k_{t} &= o_{t} \odot \text{tanh}(c_{t})
    \end{align}
Lastly, $k_{t}$ is used by the revision policy to compute the action $a_{t}$:
    \begin{align}
        \pi_{\theta}(a_{t}|a_{< t}, x_{\leq t}, y_{< t}) = \sigma(\theta^{\top}k_{t} + b_{k}) \\
        a_{t} =
        \begin{cases}
        \text{\texttt{REVISE}} ,& \text{if } \sigma(\theta^{\top}k_{t} + b_{k}) \geq \tau\\
        \text{\texttt{WRITE}}  ,& \text{otherwise}
        \end{cases} \label{eq:action}
    \end{align}
where $\theta$ is a parameter vector, $b_{k}$ is the bias, and $\tau \in [0, 1]$ is a decision threshold. According to equation (\ref{eq:action}), a \texttt{REVISE} action is selected only if the policy value is greater than or equal to $\tau$; otherwise, a \texttt{WRITE} action is chosen. This threshold can be adjusted to encourage or discourage the recomputation frequency without the need to retrain the policy. Our model is equal to an RNN when $\tau = 1$ (never recompute), and becomes a restart-incremental Transformer when $\tau = 0$ (always recompute).

\subsection{Incremental Inference Mechanism}
Using the policy, \textsc{Tapir} predicts when to perform a recomputation. Assume that an input token $x_{t}$ is fed to the RNN component to obtain $y_{t}$. The controller then reads $x_{t}$, $h_{t}$, and $\Gamma^{p}$ to compute $a_{t}$. If a \texttt{REVISE} action is emitted, the input buffer (containing all available inputs so far) will be passed to the reviser to yield the recomputed outputs. When this happens, both $z$ and $\varphi$ stored in the caches also need to be updated to reflect the effect of the recomputation. The recomputation of past $z$ and $\varphi$ will occur simultaneously with the computation of $z$ and $\varphi$ for the current time step to update $\Gamma^{z}$ and $\Gamma^{p}$ using the recomputed outputs. If a \texttt{WRITE} action is emitted, we take $y_{t}$ to be the current output and continue to process the next token. The content of $\Gamma^{z}$ and $\Gamma^{p}$ are also updated for the current step. The cache $\Gamma^{h}$ is always updated regardless of which action the policy takes. See algorithm in the Appendix.

Let us use Figure \ref{fig:inc-proc-example} and $\tau=0.5$ as a constructed example. At $t=1$, the incremental processor consumes the token $the$, updates its hidden state and predicts the POS-tag \texttt{det}. The controller predicts that the probability for recomputation is \textit{e.g.}\ 0.3. Since it is lower than $\tau$, \texttt{det} gets written to the output buffer, the memory is updated and the current step is finished. A similar decision happens at $t=2$ and \textit{alert} is classified as \texttt{noun}. At $t=3$, however, the controller predicts that a \texttt{REVISE} action should occur after the input \textit{citizens}. That triggers the reviser, which takes \textit{the alert citizens} as input and returns \texttt{det adj noun}. The output buffer gets overwritten with this new hypothesis and the caches are recomputed to accommodate the new state. This dynamics continues until the end of the sentence.

\subsection{Training}

Jointly training all components of such a two-pass model from scratch can be unstable \citep{sainath2019twopass}, so we opt for a two-step training process:

\begin{enumerate}
    \item \label{train1} Train only the reviser using cross entropy loss.
    \item \label{train2} Train the incremental processor and the controller together with a combined loss:
    \begin{align}
        \mathcal{L} = CE(y^{\text{gold}}, y) + BCE(a^{\text{LT}}, a) \label{eq:train}
    \end{align}
    where $y^{\text{gold}}$ is the expected output and $a^{\text{LT}}$ is the expected action.
\end{enumerate}

\section{Supervision Signal for Revision}
\label{sec:rev-signal}
During incremental sentence comprehension, a revision or reanalysis occurs when disambiguating material rules out the current sentence interpretation. In Figure \ref{fig:inc-proc-example}, \texttt{noun} is a valid label for \textit{suspect} at $t=6$, but \textit{person} at $t=7$ rules that analysis out, forcing a reanalysis to \texttt{adj} instead.

Training \textsc{Tapir}'s controller requires a sequence of \texttt{WRITE}/\texttt{REVISE} actions expressed as the supervision signal $a^{\text{LT}}$ in equation (\ref{eq:train}), capturing when revision happens. This signal then allows us to frame the policy learning as a supervised learning task (as in the work of \citet{zheng-etal-2019-simpler}). 

If we have the sequence of output prefix hypotheses at each step, as shown in Figure \ref{fig:inc-proc-example}, we know that the steps when revisions have occurred are $t=\{3,7\}$. We can then construct the sequence of actions we need. The first action is always \texttt{WRITE} as there is no past output to revise at this step. For $t>1$, the action can be determined by comparing the partial outputs at time step $t$ (excluding $y_{t}$) against the partial outputs at time step $t-1$. If no edits occur, then the partial outputs after processing $x_{t}$ should not change, and a \texttt{WRITE} action is appended to the sequence. If any edits occur, we append a \texttt{REVISE} action instead.

Intermediate human judgements about when to revise are not available, so we need to retrieve that from a model. 
It is possible obtain this information from a restart-incremental Transformer, by comparing how the prefix at $t$ differs from prefix at $t-1$. However, as shown by \citet{kahardipraja-etal-2021-towards}, the signal captured using this approach may lack incremental quality due to the missing recurrence mechanism. Using a recurrent model is advisable here, as it can capture order and hierarchical structure in sentences, which is apparently hard for Transformers \citep{tran-etal-2018-importance, hahn-2020-theoretical, sun-lu-2022-implicit}. But it is difficult to retrieve this signal using vanilla RNNs because its recurrence only allows a unidirectional information flow, which prevents a backward update of past outputs. 

Therefore, we opt for the Linear Transformer (LT) \citep{pmlr-v119-katharopoulos20a}, which can be viewed both as a Transformer and as an RNN.\footnote{For a detailed proof of why LT is more suitable, see the Appendix.} To generate the action sequences, we first train the action generator LT with causal mask to mimic an RNN training. Afterwards, it is deployed under restart-incrementality on the same set used for training with the mask removed. We collect the sequence of partial prefixes for all sentences and use it to derive the action sequences.

\begin{figure*}[t]
\centering

\begin{subfigure}[t]{\textwidth}
   \centering
   \includegraphics[width=0.3\linewidth]{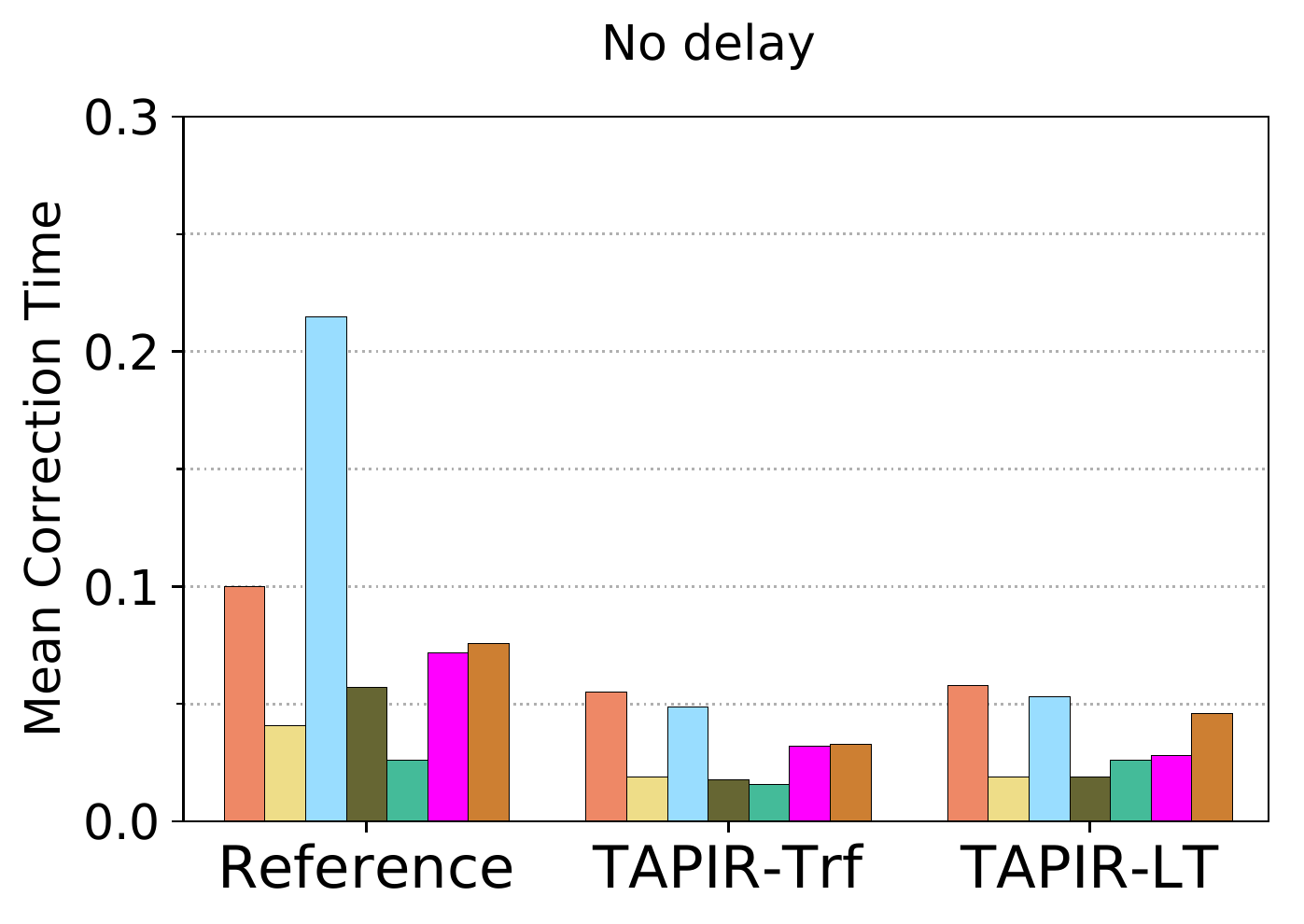}
   \hfill
   \includegraphics[width=0.50\linewidth, trim={1.75cm -1.75cm -1.75cm 0cm}]{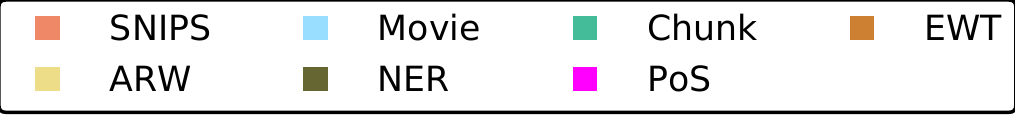}
   
\end{subfigure}
\vspace*{.1cm}
\begin{subfigure}[t]{\textwidth}
   \centering
   \includegraphics[width=0.3\linewidth]{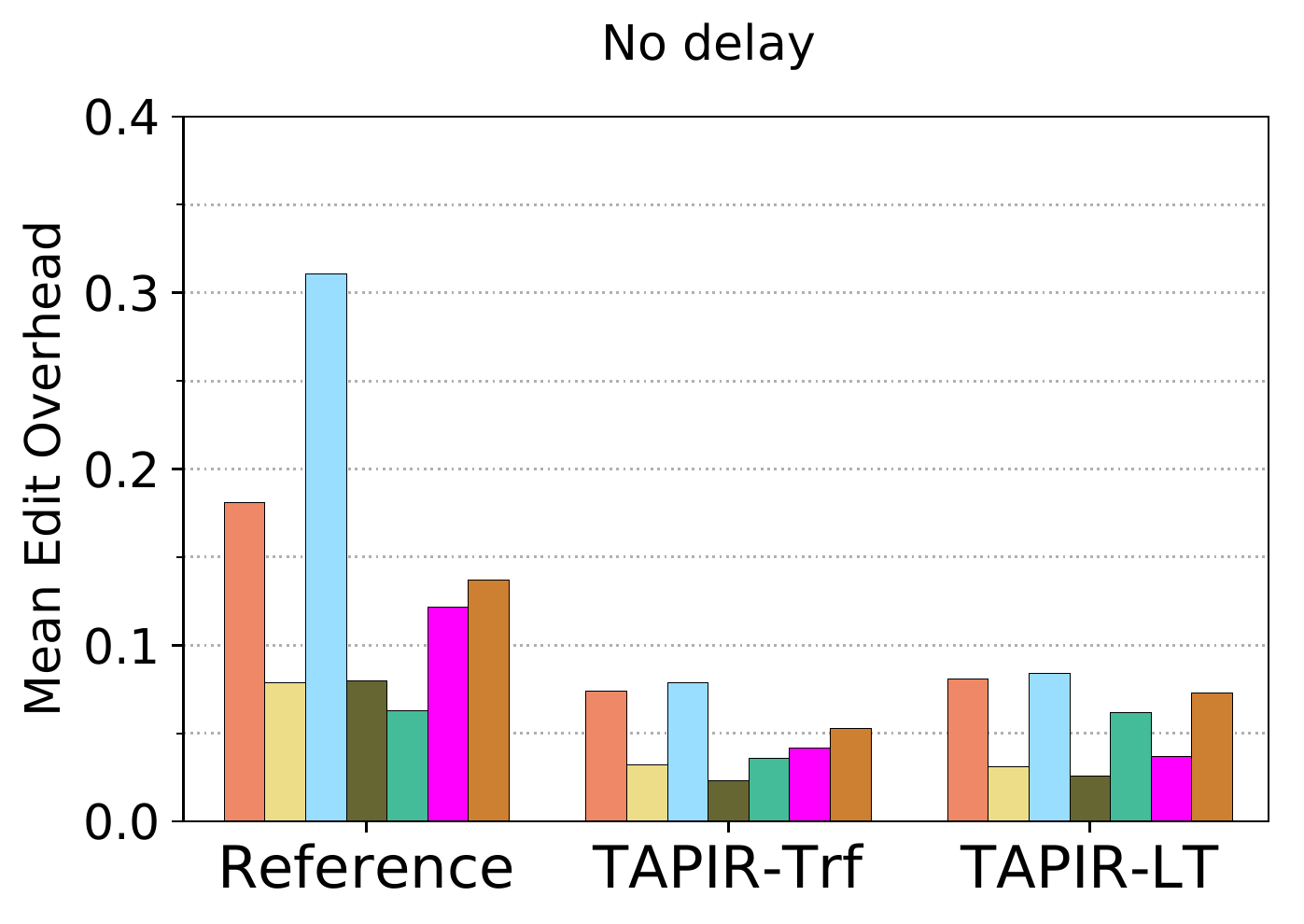}
    \hfill
  \includegraphics[width=0.3\linewidth]{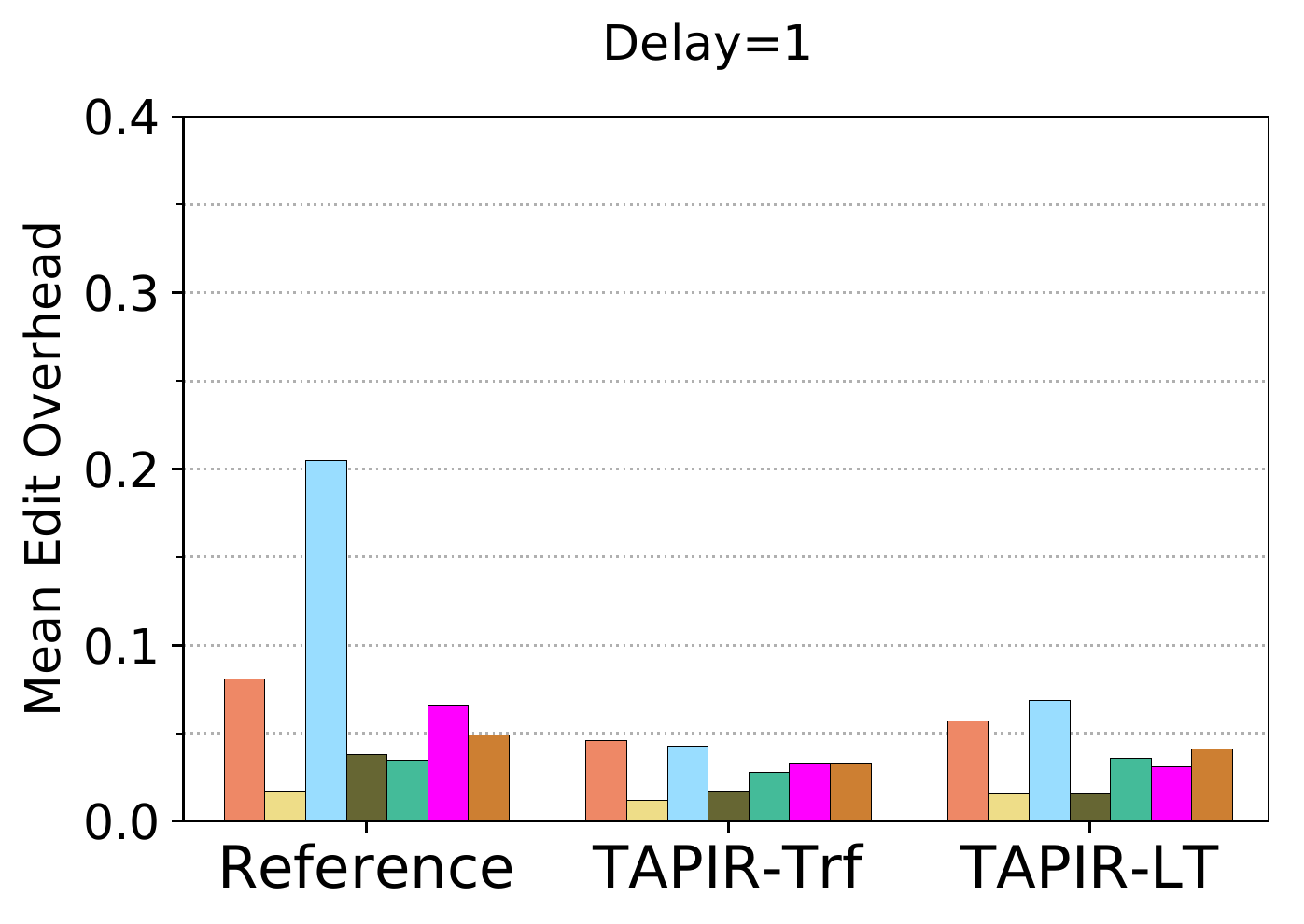}
   \hfill
   \includegraphics[width=0.3\linewidth]{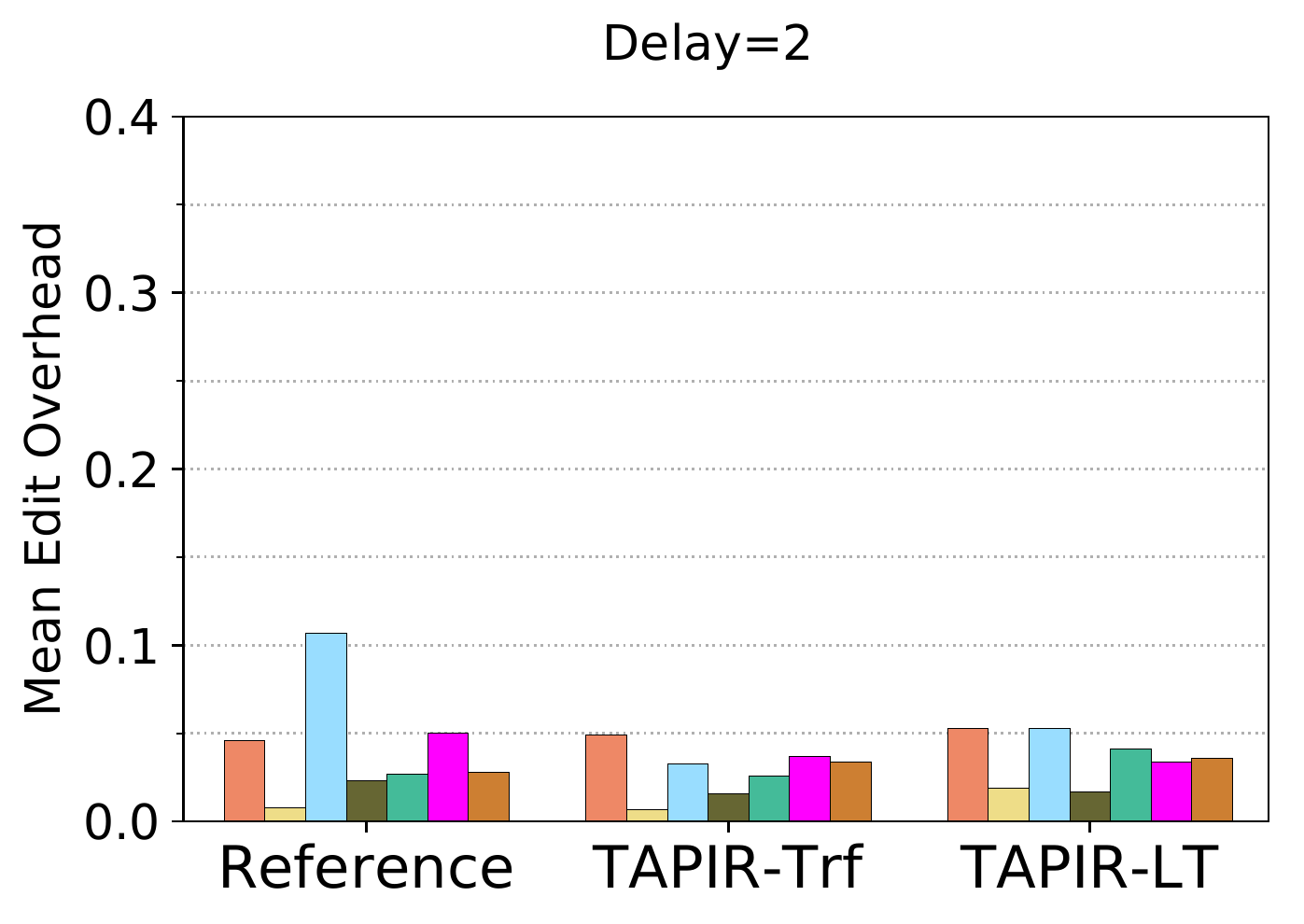}

\end{subfigure}
\vspace*{.1cm}
\begin{subfigure}[t]{\textwidth}
   \centering
   \includegraphics[width=0.3\linewidth]{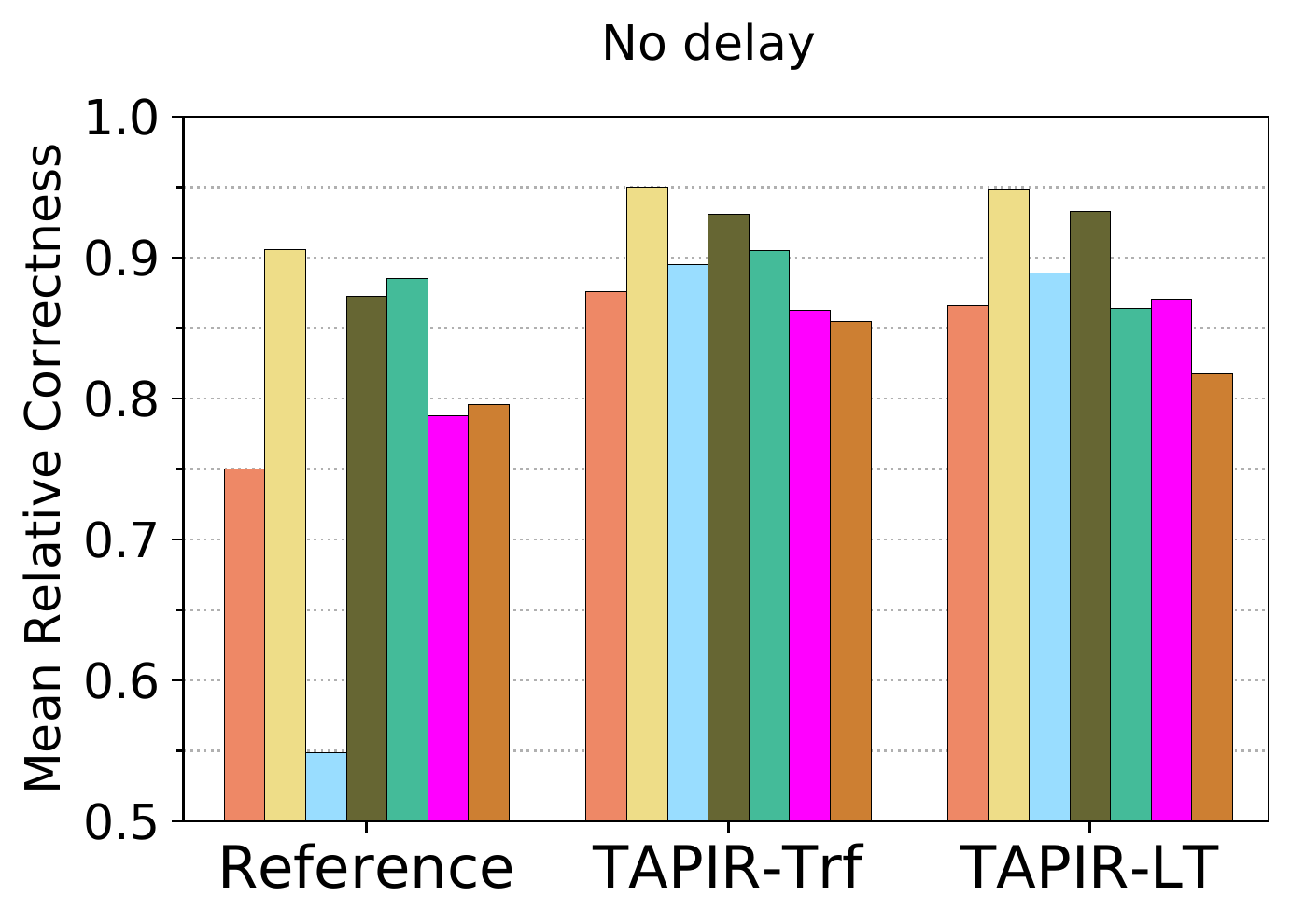}
    \hfill
  \includegraphics[width=0.3\linewidth]{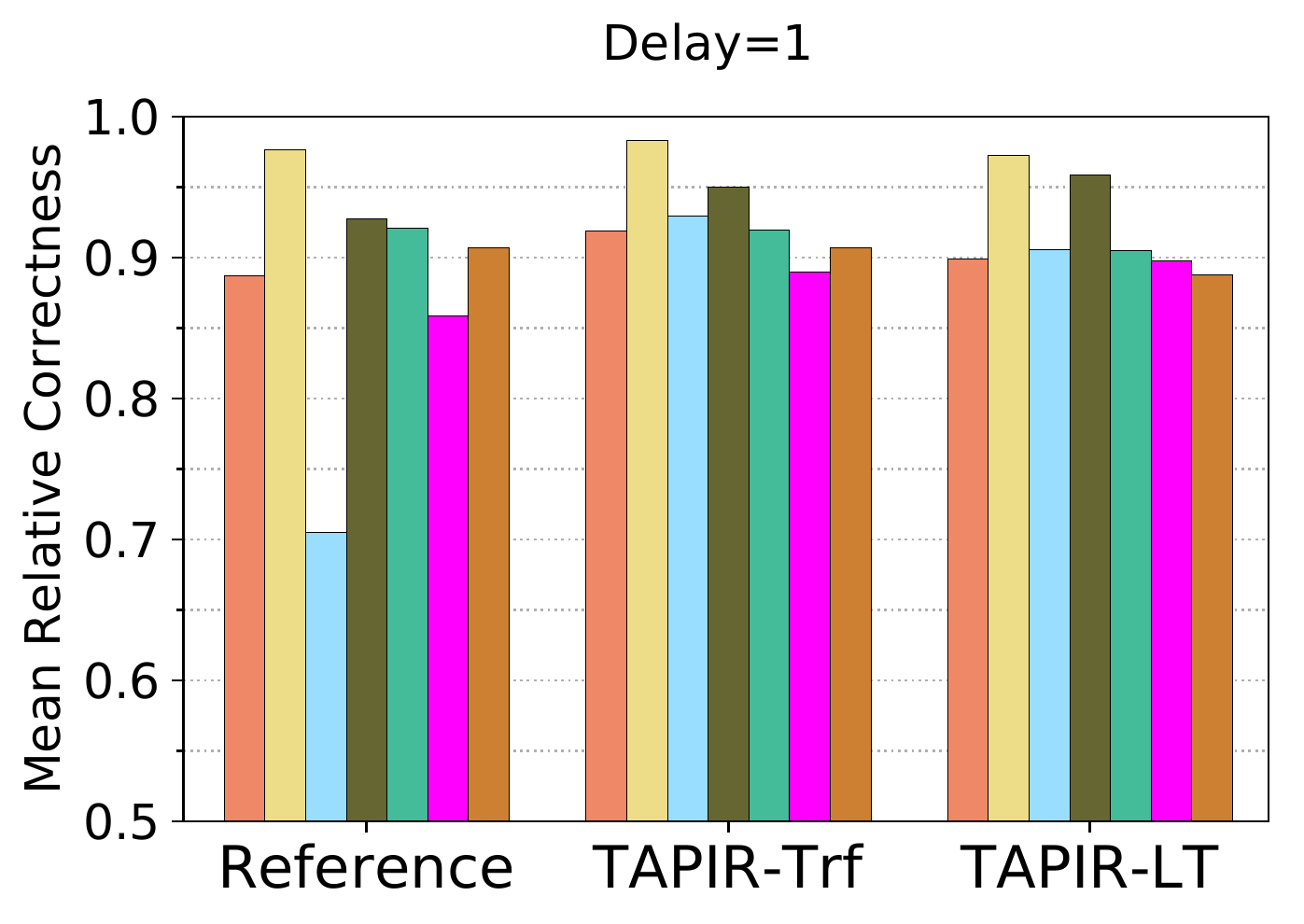}
   \hfill
   \includegraphics[width=0.3\linewidth]{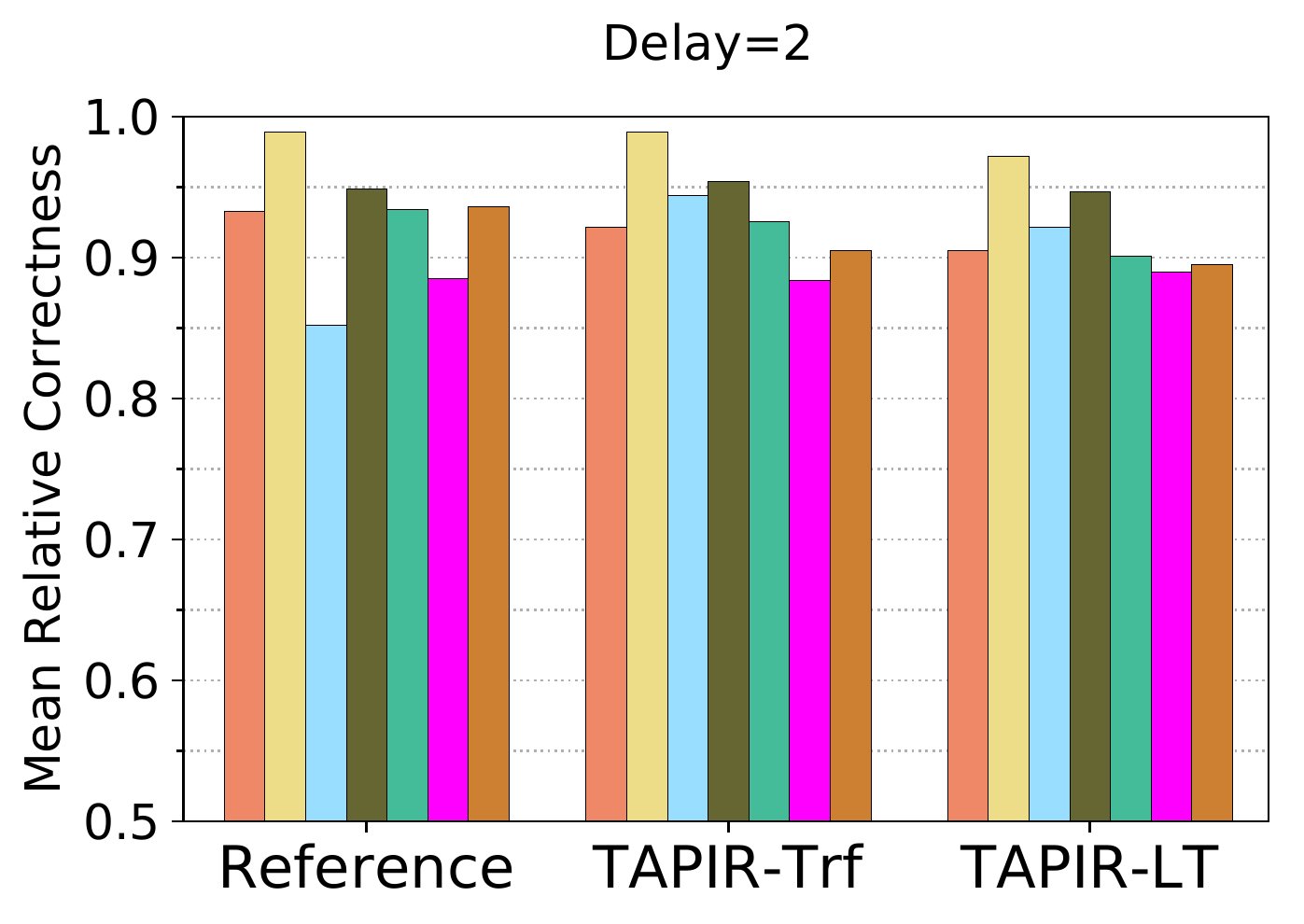}
    
\end{subfigure}
  \caption{Incremental evaluation of the models on test sets. Edit Overhead, Correction Time Score and Relative Correctness $\in [0, 1]$. Lower is better for EO and CT, while higher is better for RC. \textsc{Tapir} is better compared to the reference model for the non-delayed case (output prefixes are often correct and stable). The delay strategy of one lookahead token is beneficial.}
  
  \label{fig:incrementalmetrics} 
\end{figure*}

\section{Experiments}
\noindent\textbf{Datasets}. We evaluate \textsc{Tapir} on four tasks in English, for NLU and task-oriented dialogue, using seven sequence labelling datasets: 

\begin{itemize}
    \item[$\vartriangleright$] Slot Filling: SNIPS \citep{coucke2018snips}; Alarm, Reminder \& Weather \citep{schuster-etal-2019-cross-lingual} and MIT Movie \citep{mitrestaurantmovie}.
    \item[$\vartriangleright$] PoS Tagging: CoNLL-2003 \citep{tjong-kim-sang-de-meulder-2003-introduction} and UD-EWT \citep{silveira-etal-2014-gold}.
    \item[$\vartriangleright$] Named-Entity Recognition (CoNLL-2003). 
    \item[$\vartriangleright$] Chunking (CoNLL-2003). 
\end{itemize}

Table \ref{tab:actions-dist-train} shows the distribution of generated actions in the final training set for each task. Further details regarding the datasets and generated action sequences are available in the Appendix.

\begin{table}[h] \small
\centering
\begin{tabular}{l c c} 
\toprule
\multicolumn{1}{l}{Tasks}  & \multicolumn{1}{l}{\texttt{WRITE}} & \multicolumn{1}{l}{\texttt{REVISE}}\\
\midrule
SNIPS & 0.777 & 0.223 \\
ARW & 0.811 & 0.189 \\
Movie & 0.765 & 0.235 \\
NER & 0.895 & 0.105 \\
Chunk & 0.687 & 0.313 \\
PoS & 0.769 & 0.231 \\
EWT & 0.712 & 0.288 \\
\bottomrule
\end{tabular}
\caption{Distribution of generated actions (train+val).}
\label{tab:actions-dist-train}
\end{table}

\noindent\textbf{Evaluation}. An ideal incremental model deployed in real-time settings should (i) exhibit good incremental behaviour, \textit{i.e.}\ produce correct and stable partial hypotheses and timely recover from its mistakes; (ii) be efficient for inference by delivering responses without wasting computational resources; and (iii) not come with the cost of a negative impact on the non-incremental performance, \textit{i.e.}\ produce correct final outputs. Achieving all at the same time may be hard, so trade-offs can be necessary. 
 
 We evaluate \textsc{Tapir} on these three relevant dimensions. For (i), we use similarity and diachronic metrics\footnote{This metric can also be used for incremental evaluation involving frame semantics. See \citet{DBLP:conf/interspeech/AttererBS09} for details.} proposed by \citet{baumannetal2011} and adapted in \citet{madureira-schlangen-2020-incremental}: \emph{edit overhead} (EO, the proportion of unnecessary edits over all edits), \emph{correction time score} (CT, the average proportion of time steps required for an output increment to settle down), and \emph{relative correctness} (RC, the proportion of output prefixes that match with the final output). Aspect (ii) is analysed by benchmarking the incremental inference speed. For (iii), we use the F1 score adapted for the IOB sequence labelling scheme, except for PoS tagging, which is evaluated by measuring accuracy.

Rather than trying to beat the state-of-the art results, we focus on analysing the incremental abilities of models whose performances are high enough for our purposes. As a reference model, we use a Transformer encoder applied in a restart-incremental fashion, which implicitly performs revision at every step. We follow \citet{baumannetal2011} and \citet{madureira-schlangen-2020-incremental} by evaluating partial outputs with respect to the final output, to separate between incremental and non-incremental performance. 

\noindent\textbf{Delay strategy}. To inspect the effect of right context on the model's performance, we use the delay strategy \citep{baumannetal2011} with a lookahead window of size 1 and 2, computing a delayed version of EO and RC \citep{madureira-schlangen-2020-incremental}. The output for the reference model is delayed only during inference, as in \citet{madureira-schlangen-2020-incremental}. For \textsc{Tapir}, the same treatment would not be possible as it contains an RNN that must be able to recognise the output delay. Thus, we follow the approach of \citet{pmlr-v119-turek20a}: During training and inference, the label for input $x_{t}$ is expected at time step $t+d$, where $d$ is the delay.

\noindent\textbf{Implementation}. For the reviser component, we choose Transformer (Trf) and Linear Transformer (LT) encoders trained with full attention.\footnote{We previously tried to train both revisers with causal mask to make revision more robust as it can occur before the input is complete, however our preliminary results show that training without mask yields better results.} The reference model is trained with cross entropy loss similar to the reviser. All models are trained with the AdamW optimiser \citep{loshchilov2018decoupled}. We use 300-D GloVe embeddings \citep{pennington-etal-2014-glove}, which, for the reference model and the reviser, are passed through an additional linear projection layer. The probability threshold $\tau$ is set to 0.5. We report results for a single run with the best hyperparameter configuration. See Appendix for details about the set-up and experiments.

\section{Results and Analysis}

\noindent\textbf{Incremental}. Figure \ref{fig:incrementalmetrics} depicts the incremental evaluation results. For the no-delay case, \textsc{Tapir} performs better compared to the reference model. We also observe that the delay strategy helps improve the  metrics. It improves the results for \textsc{Tapir}, in general, but a longer delay does not always yield a better incremental performance. We suspect this happens for two possible reasons: First, if we consider the case where the delay is 1, \textsc{TAPIR} has already achieved relatively low EO ($< 0.1$) and high RC ($ > 0.85$). This, combined with its non-monotonic behaviour, might make it harder to further improve on both incremental metrics, even if a longer delay is allowed. Second, a longer delay means that our model needs to wait longer before producing an output. In the meantime, it still has to process incoming tokens, which might cause some difficulty in learning the relation between the input and its corresponding delayed output. As a consequence, we have mixed results when comparing EO and RC for the delayed version of the reference model and \textsc{Tapir}. Their differences are, however, very small. \textsc{Tapir} achieves low EO and CT score, which indicates that the partial output is stable and settles down quickly. RC is also high, which shows that, most of the time, the partial outputs are correct prefixes of the final, non-incremental output and would be useful for downstream processing. 

\noindent\textbf{Benchmark}. Table \ref{table:benchmark} shows that \textsc{Tapir} is considerably faster compared to the reference model in incremental settings, as it offers, on average, $\sim$4.5$\times$ speed-up in terms of sequences per second.\footnote{We use the same models as in Table \ref{table:nonincremental} and Figure \ref{fig:incrementalmetrics} for benchmarking, as the policy affects the inference speed.} 

\begin{table}[h!] \small
\captionsetup{singlelinecheck = false, justification=justified}
\setlength\tabcolsep{5pt}
\begin{center}
\begin{tabular}{l c c c}
\toprule
 \multicolumn{1}{l}{Tasks} & \multicolumn{1}{c}{Ref.} & \multicolumn{1}{c}{\textsc{Tapir}-Trf} & \multicolumn{1}{c}{\textsc{Tapir}-LT} \\
\midrule
SNIPS     & 1.103 & \hphantom{0}4.958 (4.50$\times$) &  \bf 8.983 (8.15$\times$)   \\
ARW    & 2.339 &  \bf \hphantom{0}8.734 (3.73$\times$) & 5.959 (2.55$\times$)  \\
Movie   	  & 0.927 &  \bf \hphantom{0}3.520 (3.80$\times$) & 3.432 (3.70$\times$)   \\
NER 		  & 0.675 & \hphantom{0}4.465 (6.62$\times$) &  \bf 4.502 (6.67$\times$)    \\
Chunk   & 0.688 &  \bf \hphantom{0}2.714 (3.95$\times$) & 1.912 (2.78$\times$)   \\
PoS   & 0.672 & \hphantom{0}4.111 (6.12$\times$) &  \bf \hphantom{0}7.400 (11.01$\times$)   \\
EWT  & 0.819 &  \bf \hphantom{0}3.659 (4.47$\times$) & 3.122 (3.81$\times$)   \\
\midrule
Average & 1.032 & \hphantom{0}4.594 (4.45$\times$) & \textbf{5.044} (\textbf{4.89}$\times$) \\
\bottomrule
\end{tabular}
\end{center}
\caption{Comparison of incremental inference speed on test sets. \textsc{Tapir} is $\sim$4.5$\times$ faster compared to the reference model. All results are in sentences/sec.}
\label{table:benchmark}
\end{table}

\noindent\textbf{Non-Incremental}. The performance of the restart-incremental reference model and our model on full sentences is shown in Table \ref{table:nonincremental}. The results of \textsc{Tapir}, in particular with the Transformer reviser (\textsc{Tapir}-Trf), are roughly comparable to the reference model, with only modest differences (0.96\% -- 4.12\%). \textsc{Tapir}-Trf performs slightly better than \textsc{Tapir}-LT. This is possibly due to the approximation of softmax attention in LT, which leads to degradation in the output quality. Furthermore, we see that delay of 1 or 2 tokens for \textsc{Tapir} is generally beneficial.\footnote{As \textsc{Tapir} primarily processes sequences left-to-right and only recomputes upon emission of a \texttt{REVISE} action, the delay strategy helps to provide more information when making a decision, unlike the reference model that already has access to full sequences.} Note that we do not force a \texttt{REVISE} action at the final time step to examine the effect of the learned policy on \textsc{Tapir}'s performance, although that would be a strategy to achieve the same non-incremental performance as the reference model.

\begin{table}[ht] 
{\footnotesize
\captionsetup{singlelinecheck = false, justification=justified}
\setlength\tabcolsep{3.5pt}
\label{turns}
\begin{tabular}{l @{\;\;} c @{\;\;\;} c c @{\;\;} c @{\;\;} c @{\;\;} c @{\;\;} c }
\toprule
 &  & \multicolumn{3}{c}{\textsc{Tapir}-Trf} & \multicolumn{3}{c}{\textsc{Tapir}-LT} \\
    \cmidrule(lr){3-5} \cmidrule(lr){6-8}
 Tasks     & Ref. &  - & D1 & D2 & - & D1 & D2\\
\midrule
SNIPS     & \bf 91.05 & 88.57 & \cellcolor{lightgray!40}{90.47} & 89.45 & 85.95 & 88.07 & 87.28 \\
ARW    & \bf 95.63 & 93.35 & \cellcolor{lightgray!40}95.17 & 95.15 & 92.84 & 93.65 & 94.50  \\
Movie    & \bf 83.98 & 82.85 & \cellcolor{lightgray!40}83.26 & 82.95 & 81.40 & 83.16  & 82.21 \\
NER    & \bf 78.25 & 74.13 & 76.85 & \cellcolor{lightgray!40}78.04 & 73.12 & 73.79 & 75.75  \\
Chunk    & \bf 88.35 & 86.85 & 87.48 & \cellcolor{lightgray!40}87.52 & 85.03 & 86.43 & 85.79  \\
\midrule
PoS  & \bf 92.28 & 91.32 & 91.35 & \cellcolor{lightgray!40}91.49 & 90.90 & 90.83 & 90.65   \\
EWT  & \bf 92.14 & 90.84 & \cellcolor{lightgray!40}92.00 & 91.95 & 90.20 & 91.34 & 90.93  \\

\bottomrule
\end{tabular}

\caption{Non-incremental performance of the models on test sets (first group is F1, second group is accuracy). $D=$ delay. The performance of \textsc{Tapir} is roughly comparable to the reference model.}
\label{table:nonincremental}
}

\end{table}

\subsection{Detailed Analysis}

In the next paragraphs, we assess \textsc{Tapir}-Trf on aspects beyond the basic evaluation metrics. 

\begin{figure}[h]
	\centering
    \includegraphics[width=0.9\columnwidth,trim={0cm 0cm 1cm 0cm}]{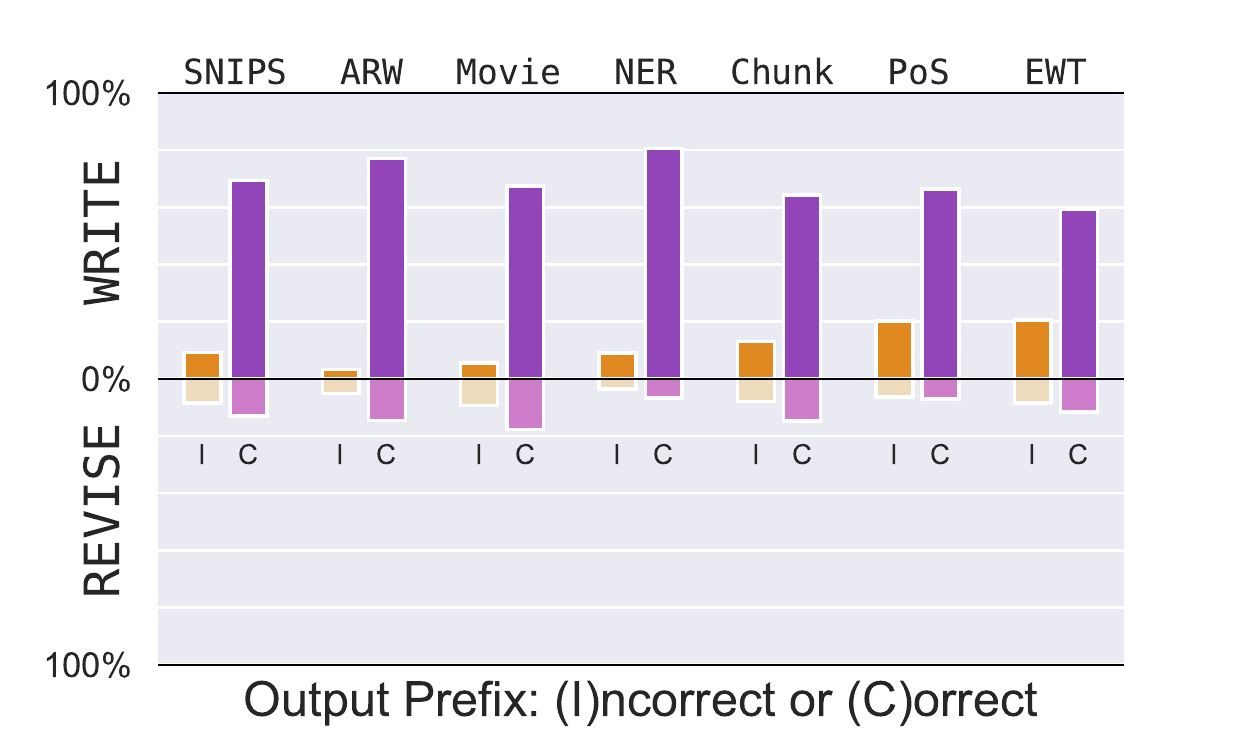}
	\caption{Distribution of actions and output prefixes by dataset. Most of the actions are \texttt{WRITE} and most of the partial prefixes which are correct do not get unnecessarily revised. Incorrect prefixes cannot always be immediately detected, as expected. Part of the \texttt{REVISE} actions are dispensable, but in a much lower frequency than in the restart-incremental paradigm.
 }
	\label{fig:policy-details}
\end{figure}

\noindent\textbf{Policy Effectiveness}. Figure \ref{fig:policy-details} shows the distributions of actions and states of the output prefixes. Here, a prefix is considered correct if all its labels match the final output, and incorrect otherwise. We start by noticing that most of the actions are \texttt{WRITE}, and among them, very few occur when the prefix is incorrect. \textsc{Tapir} is thus good at recognising states where recomputation is not required, supporting its speed advantage. A good model should avoid revising prefixes that are already correct. We see that, for all datasets, the vast majority of the correct prefixes indeed do not get revised. A utopian model would not make mistakes (and thus never need to revise) or immediately revise incorrect prefixes. In reality, this cannot be achieved, given the incremental nature of language and the long-distance dependencies. As a result, incorrect prefixes are expected to have a mixed distribution between actions, as the model needs to wait for the edit-triggering input, and our results corroborate that. Finally, among the \texttt{REVISE} actions (\textit{i.e.}\ the lighter bars in the bottom area), there is still a considerable relative number of unnecessary revisions occurring for correct prefixes. We see room for further refinement of the policy in that sense, but, in absolute numbers, the occurrence of recomputations is much lower than in the restart-incrementality paradigm, where all steps require a recomputation.

\begin{figure}[t]
	\centering
    \includegraphics[width=0.9\columnwidth, trim={2.5cm 6.75cm 2.5cm 6.5cm},clip, page=2]{acl2023_figure_example.pdf}
    \includegraphics[width=0.9\columnwidth, trim={5.5cm 7.5cm 3.3cm 5.5cm},clip, page=1]{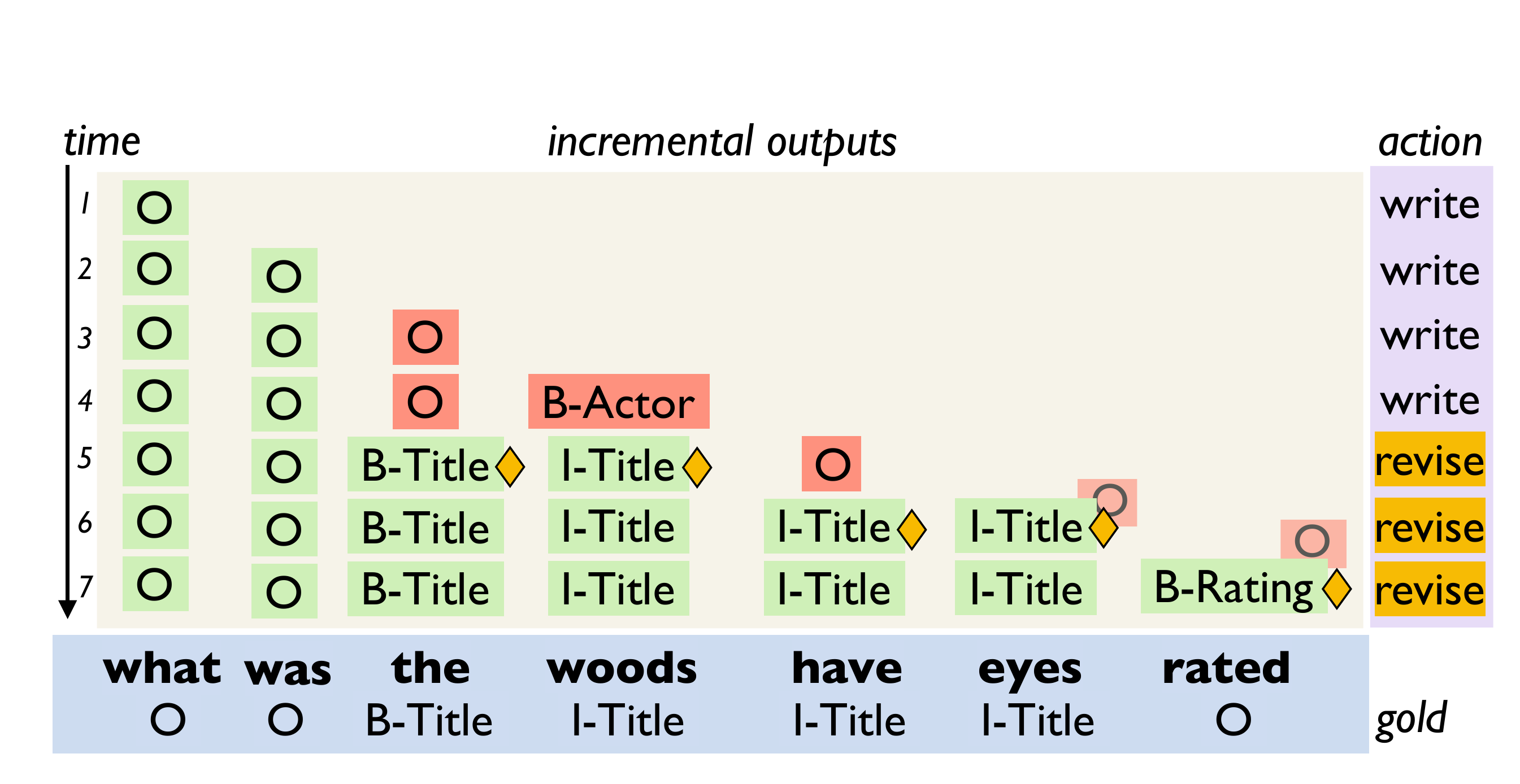}
	\caption{Examples of incremental inference (from SNIPS and Movie) for \textsc{Tapir}-Trf. Edited labels are marked by a diamond symbol, with the immediate past output at the top right corner for right-frontier edits. Red labels are incorrect with respect to the final output.}
	\label{fig:qualitative}
\end{figure}

\begin{figure*}[ht]
\centering
\begin{subfigure}[b]{0.5\textwidth}
   \centering
   \includegraphics[width=\linewidth]{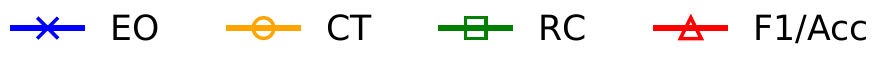}
\end{subfigure}

\begin{subfigure}[t]{\textwidth}
   \centering
   \includegraphics[width=0.3\linewidth]{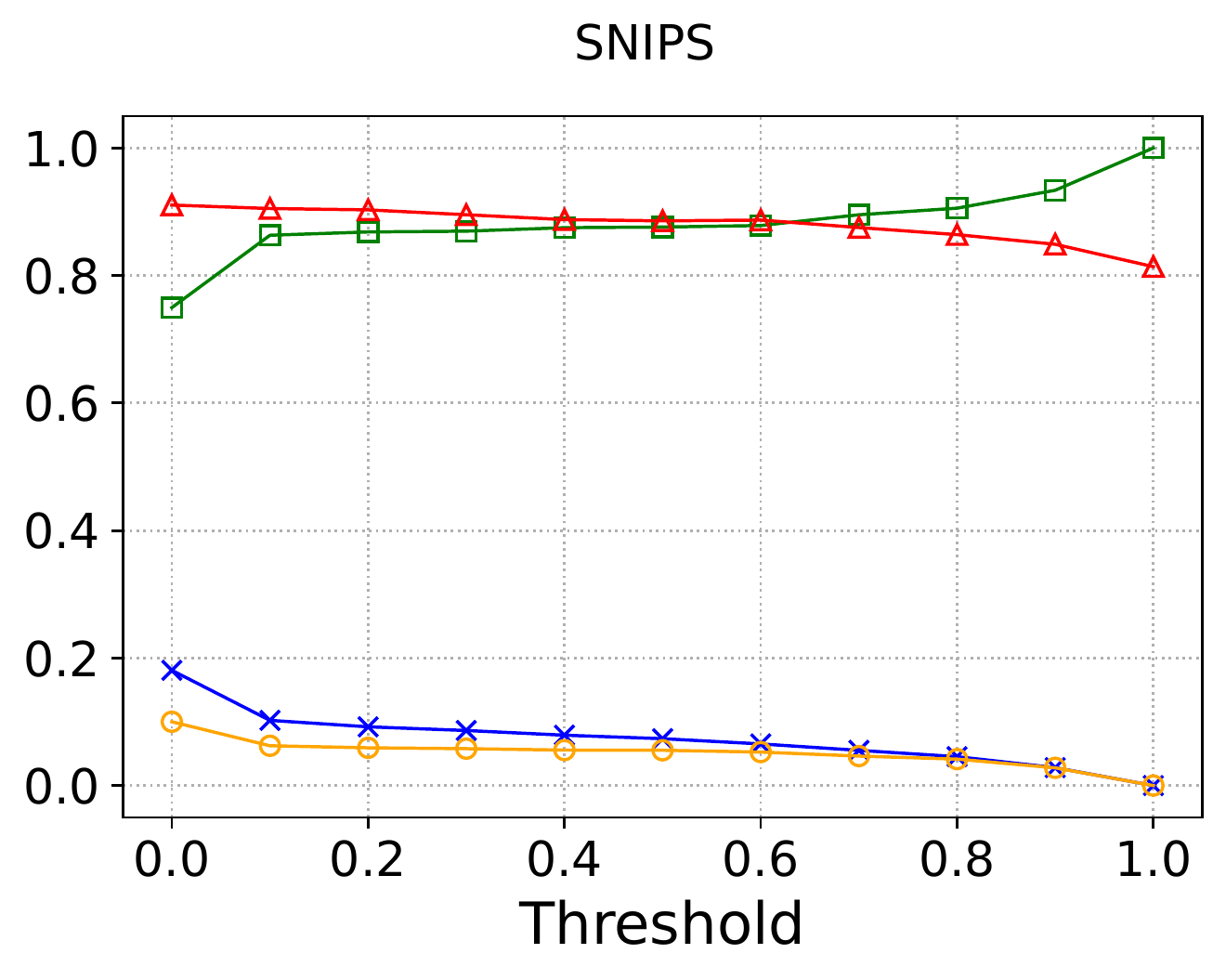}
    \hfill
  \includegraphics[width=0.3\linewidth]{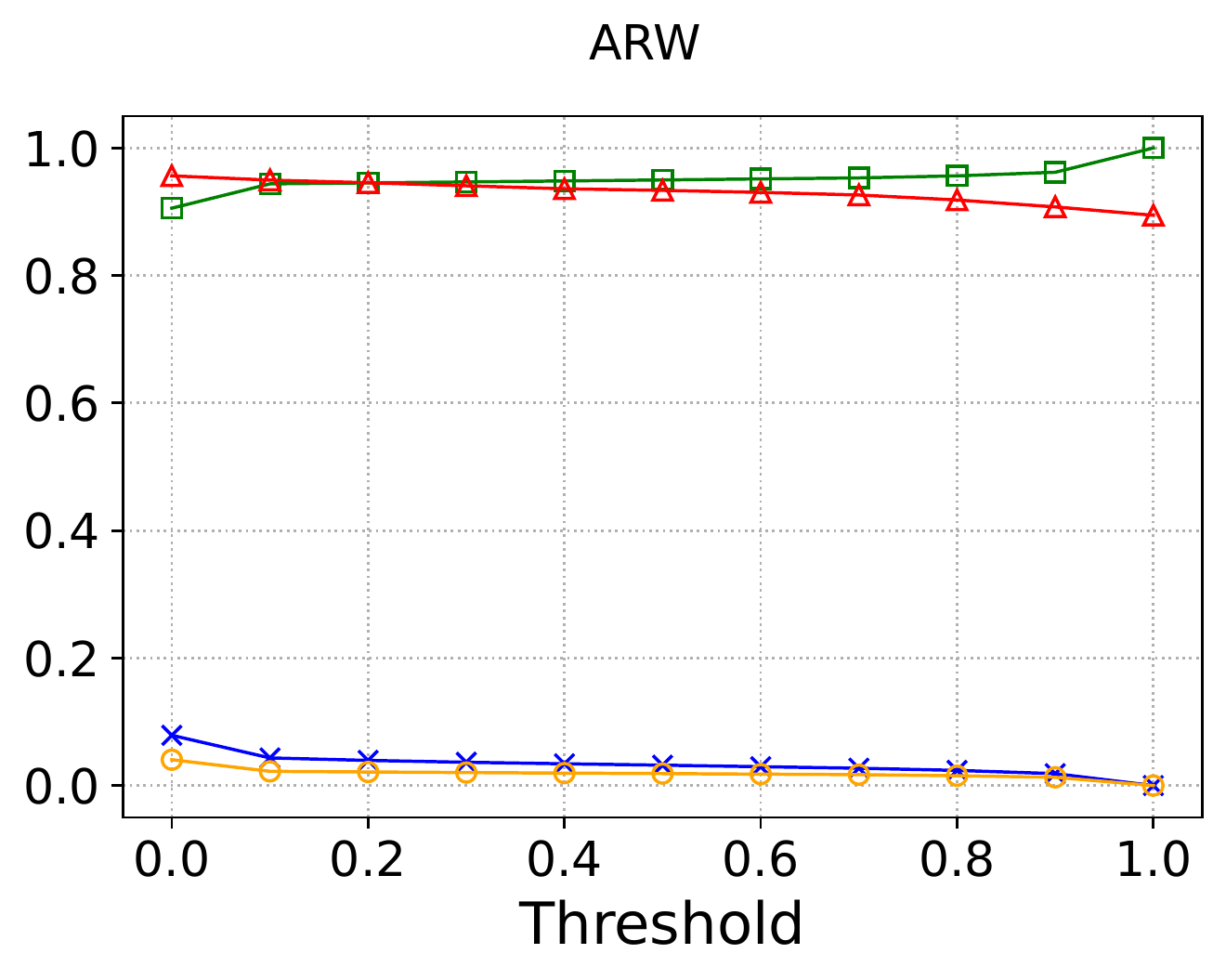}
   \hfill
   \includegraphics[width=0.3\linewidth]{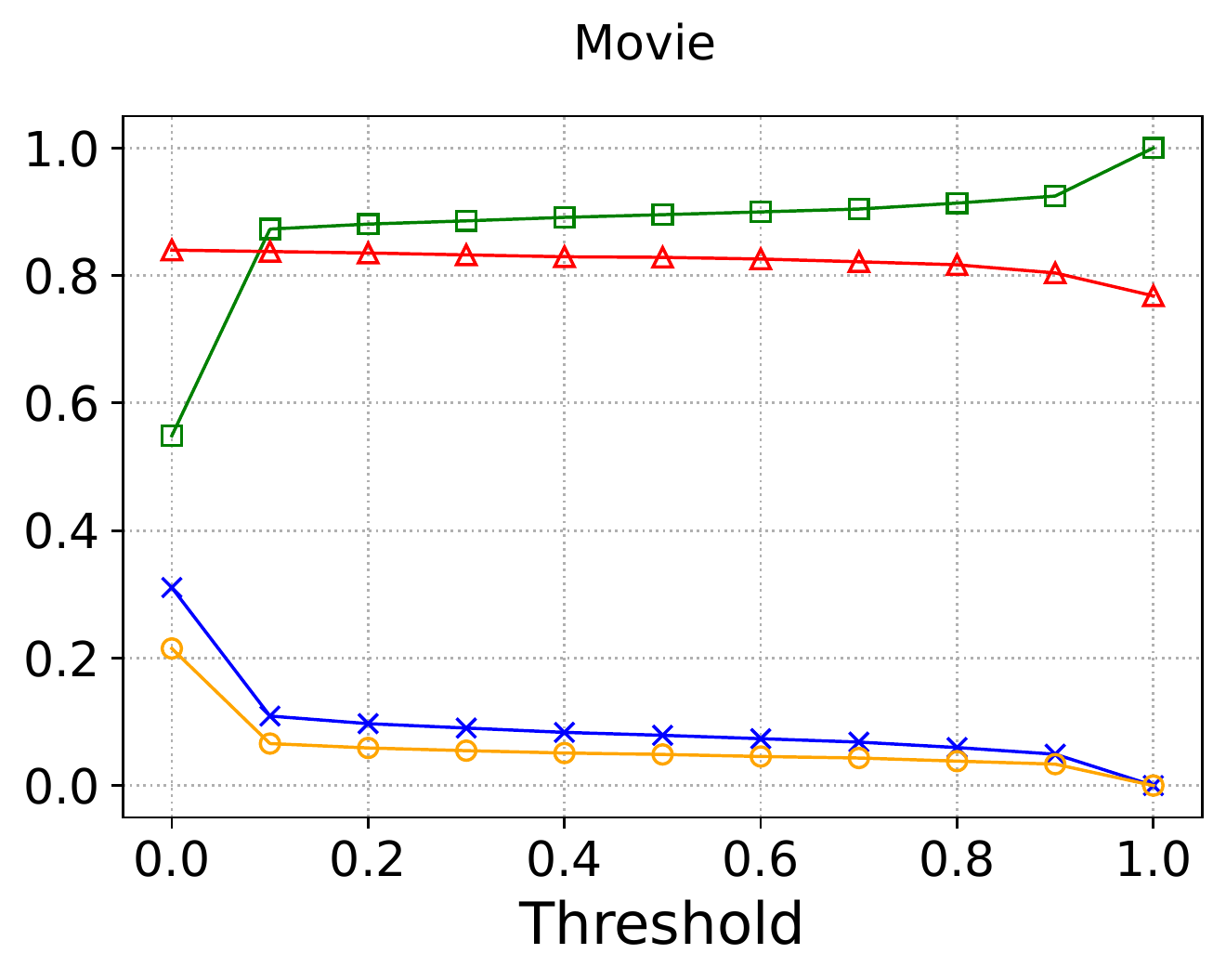}
\end{subfigure}

\begin{subfigure}[t]{\textwidth}
   \centering
   \includegraphics[width=0.3\linewidth]{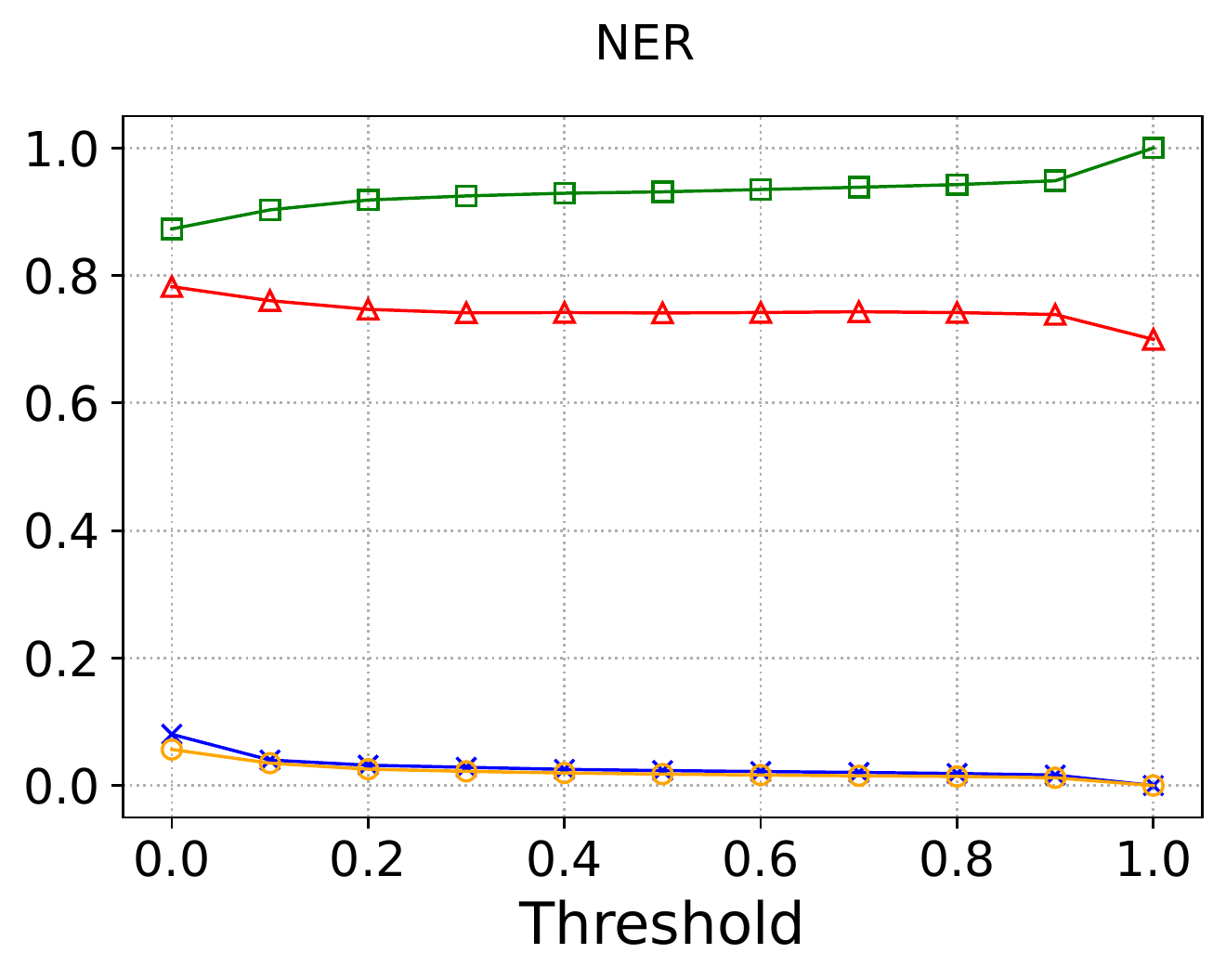}
    \hfill
  \includegraphics[width=0.3\linewidth]{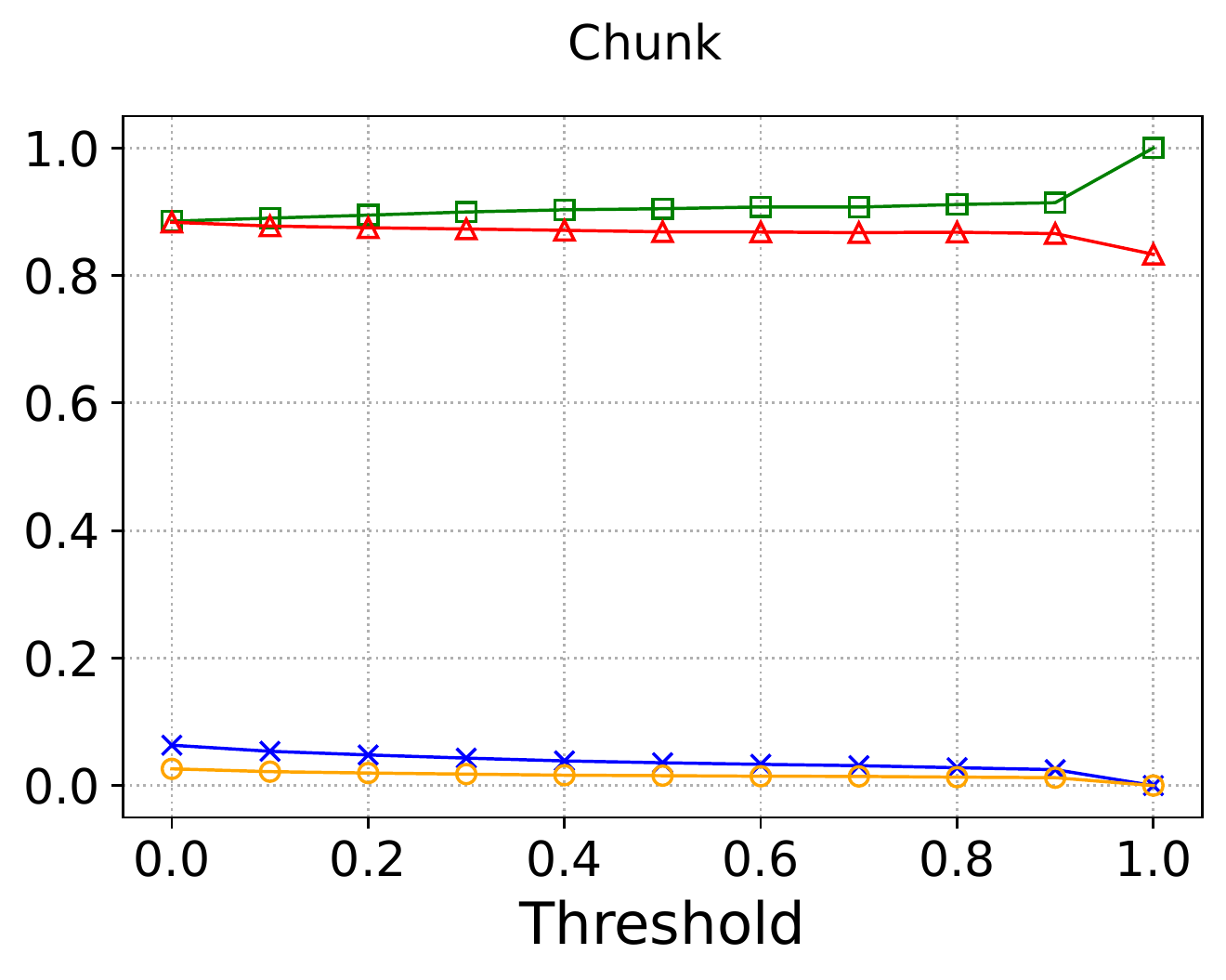}
   \hfill
   \includegraphics[width=0.3\linewidth]{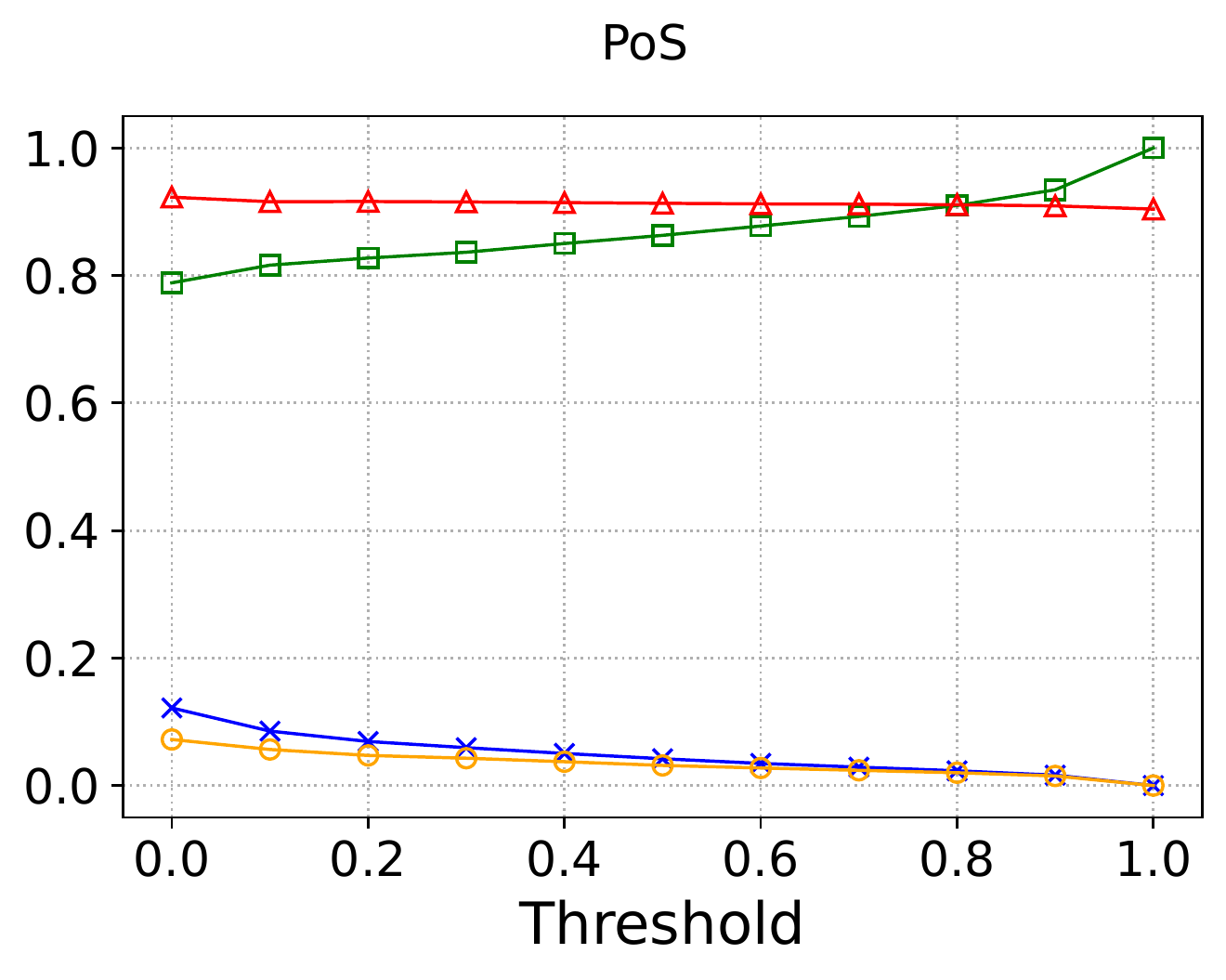}
\end{subfigure}

\begin{subfigure}[t]{\textwidth}
   \centering
   \includegraphics[width=0.3\linewidth]{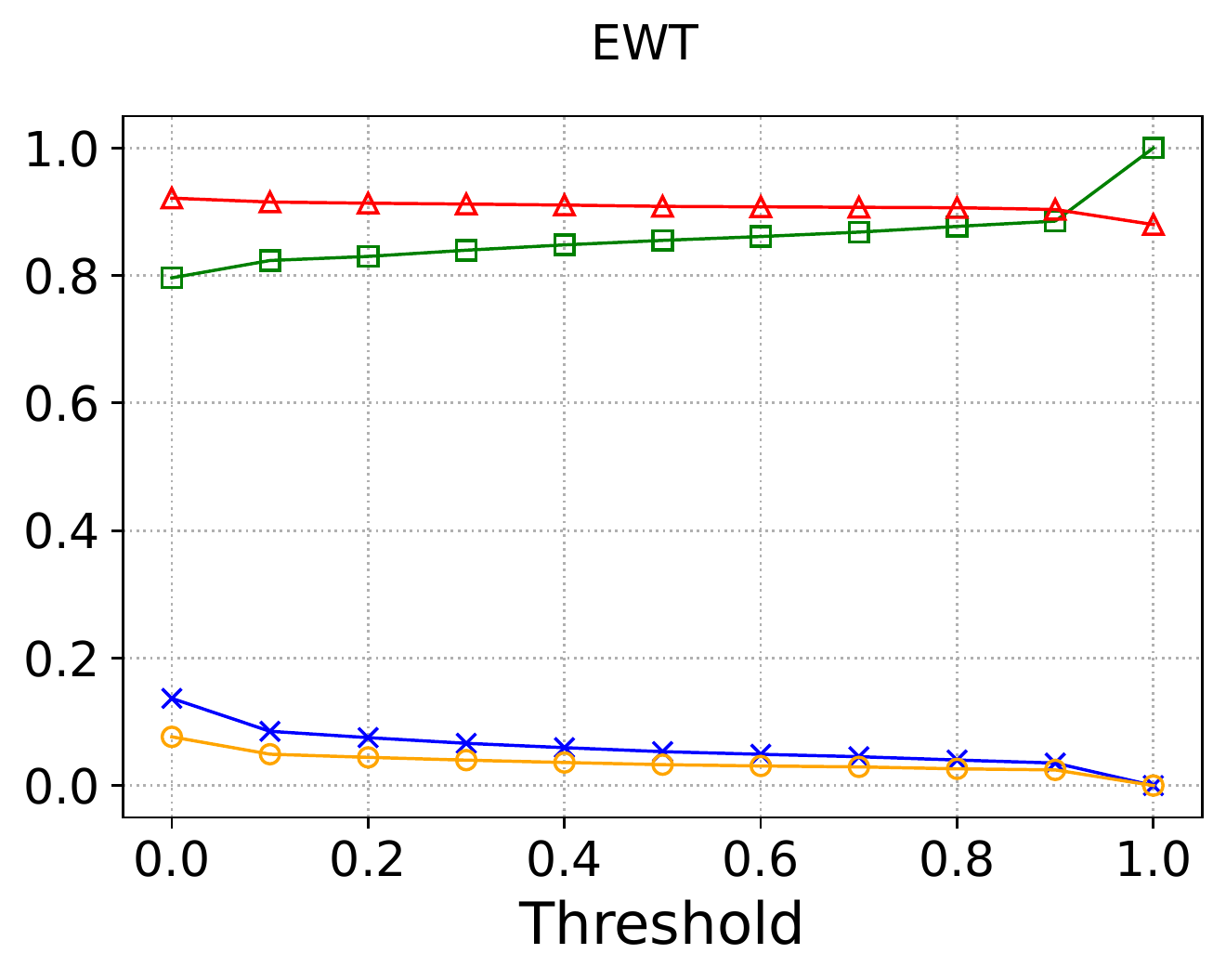}
\end{subfigure}

  \caption{Effect of the probability threshold $\tau$ on incremental and non-incremental metrics, using \textsc{Tapir}-Trf. Increasing $\tau$ leads to improvement of incremental metrics at the cost of non-incremental performance.}
  \label{fig:threshold} 
\end{figure*}

\noindent\textbf{Qualitative analysis}. Figure \ref{fig:qualitative} shows two examples of how \textsc{Tapir} behaves in incremental slot filling (more examples in the Appendix), showing that it performs critical revisions that would not be possible with a monotonic model.

At the top, the model must produce labels for unknown tokens, which is harder to perform correctly. The first \texttt{UNK} token is initially interpreted as a city at $t=6$, which is probably deemed as correct considering the available left context. The controller agrees with this, producing a \texttt{WRITE} action. However, when \textit{heritage} and the second \texttt{UNK} token have been consumed at $t=8$, the incremental processor labels them as parts of a geographic point of interest. The controller is able to notice the output inconsistency as \texttt{I-geographic\_poi} should be preceded by \texttt{B-geographic\_poi} (following the IOB scheme) and emits a \texttt{REVISE} action. As a result, the label \texttt{B-city} is correctly replaced.

In the second example, \textsc{Tapir} produces interesting interpretations. It initially considers \textit{woods} to be an actor name at $t=4$. When it reads \textit{have}, the reanalysis triggered by the controller interprets \textit{woods} as a part of a title, \textit{the woods}. The model revises its hypothesis again at $t=6$, and decides that the complete title should be \textit{the woods have eyes}. It still makes a mistake at the last time step, opting for a (wrong) revision of \texttt{O} to \texttt{B-RATING} for \textit{rated} when it should be unnecessary. 

\noindent\textbf{Effect of Threshold}. Figure \ref{fig:threshold} portrays the effect of the probability threshold $\tau$ on incremental and non-incremental metrics. As $\tau$ increases, the incremental performance improves while the non-incremental performance deteriorates. This happens as higher $\tau$ discourages recomputation and makes the model closer to an RNN. In return, it is harder for the model to revisit its past decisions. 

\section{Conclusion}

We proposed \textsc{Tapir}, a two-pass model capable of performing adaptive revision in incremental scenarios \textit{e.g.}~for dialogue and interactive systems. We also demonstrated that it is possible to obtain an incremental supervision signal using the Linear Transformer (LT), in the form of \texttt{WRITE}/\texttt{REVISE} action sequences, to guide the policy learning for adaptive revision. Results on sequence labelling tasks showed that \textsc{Tapir} has a better incremental performance than a restart-incremental Transformer, in general, while being roughly comparable to it on full sentences. The delay strategy helps to improve incremental and non-incremental metrics, although a longer delay does not always yield better results. 

The ability to revise adaptively provides our model with substantial advantages over using RNNs or restart-incremental Transformers. It can fix incorrect past outputs after observing incoming inputs, which is not possible for RNNs. Looking from the aspect of efficiency, our model is also better compared to restart-incremental Transformers as the recomputation is only performed when the need for it is detected. \textsc{Tapir} is consequently faster in terms of inference speed. 

\section*{Limitations}

In this section, we discuss some of the known limitations of our set-up, data and models.

To handle unknown words in the test sets, we replace them by a special \texttt{UNK} token which is also used to mask some tokens in the training set. The \texttt{UNK} token provides little information regarding the actual input and \textsc{Tapir} might be unable to fully utilise the token to refine its interpretation of the past output. This has a direct influence in the incremental metrics, as the model can exploit this property by using \texttt{UNK} token as a cue to emit the \texttt{REVISE} action. This strategy also introduces the extra hyperparameter of what proportion of tokens to mask.

We put effort into achieving a diverse selection of datasets in various tasks, but our analysis is limited to English. We are reporting results on the datasets for which the non-incremental versions of the model could achieve a performance high enough to allow a meaningful evaluation of their incremental performance. Tuning is still required to extend the analysis to other datasets.

Related to these two issues, we decided to use tokens as the incremental unit for processing. We follow the tokenization given by the sequence labelling datasets we use. Extending the analysis for other languages requires thus a good tokenizer, and annotated data, which may not exist. We may also inherit limitations from the datasets that we use. Although we do not include an in-depth analysis of the datasets, as our focus is on the model and not on solving the tasks themselves, they are widely used by the community and details are available in their corresponding publications.

The method we propose to retrieve the action sequences depends on the chosen model, and the grounding of the action sequences in the actual prefix outputs have a direct influence in training the controller. Therefore, the decisions made by \textsc{Tapir} rely on the quality of the underlying generated action sequences. In order to ensure that the internal representations of the action generator LT do not depend on right context, we had to restrict ourselves to a single layer variation of this model when generating the sequence of actions. It is possible that with more layers its behaviour would be different, but that would invalidate the assumptions needed for an incremental processor.

When it comes to the \textsc{Tapir} architecture, the attention scores for the controller are computed independently of temporal order and we do not explicitly model relation between cache elements. The limited cache size also means that some past information has to be discarded to accommodate incoming inputs. Although we have made efforts to incorporate them through the summary vector, this might be not ideal due to information bottleneck.

\section*{Ethics Statement}

We do not see any immediate ethical issues arising from this work, beyond those inherent to NLP which are under discussion by the community.

\section*{Acknowledgements}
We thank the anonymous reviewers for their valuable and insightful comments and suggestions. This work is partially funded by the Deutsche Forschungsgemeinschaft (DFG, German Research Foundation) -- Project ID 423217434 (Schlangen).



\bibliography{anthology,custom}
\bibliographystyle{acl_natbib}

\clearpage

\appendix

\section{Appendix}
\label{sec:appendix}
\begin{figure*}[t]
	\centering
    \begin{subfigure}[b]{0.65\textwidth}
    \centering
    \includegraphics[width=\linewidth, trim={1.5cm 6.75cm 2.5cm 6.5cm},clip, page=3]{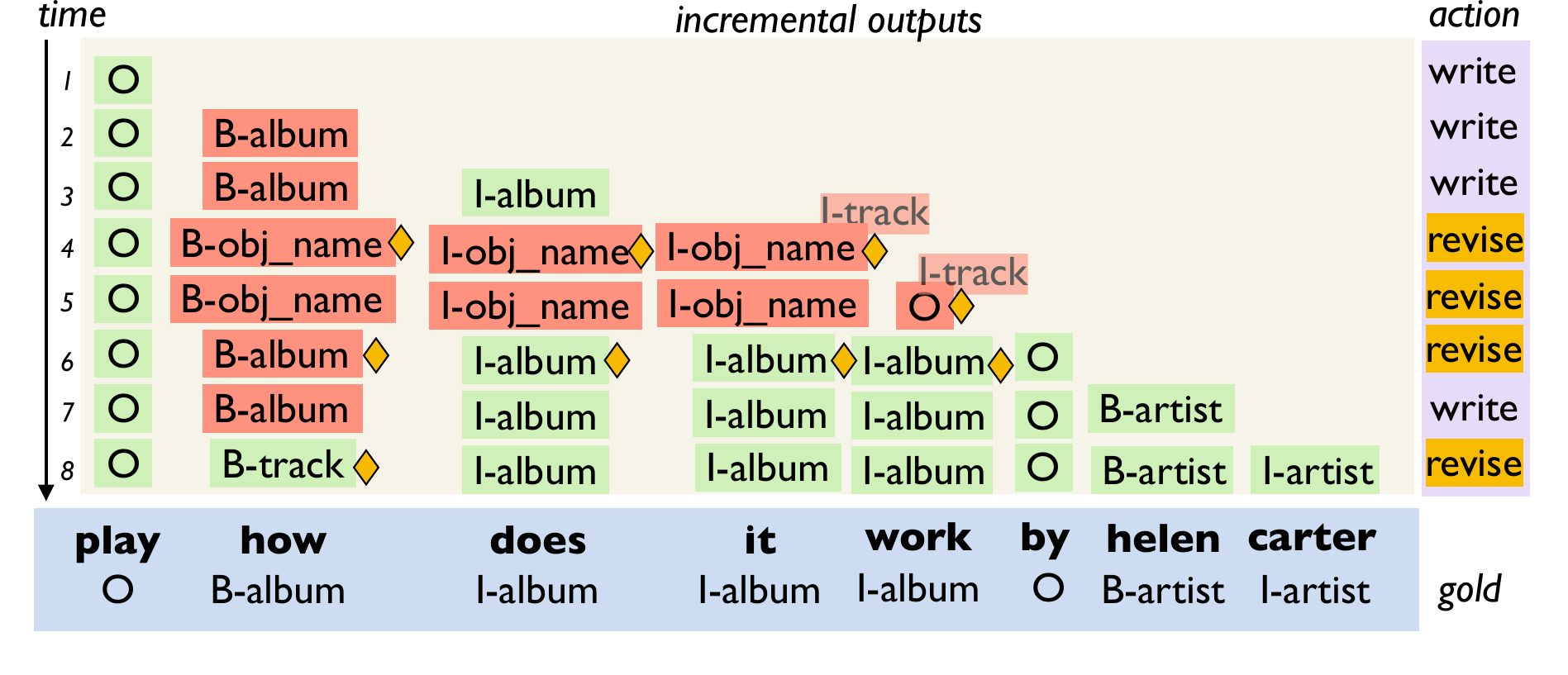}
    \end{subfigure}
    
    \begin{subfigure}[b]{0.65\textwidth}
    \centering
     \includegraphics[width=\linewidth, trim={0cm 5.5cm 0.5cm 5.5cm},clip, page=4]{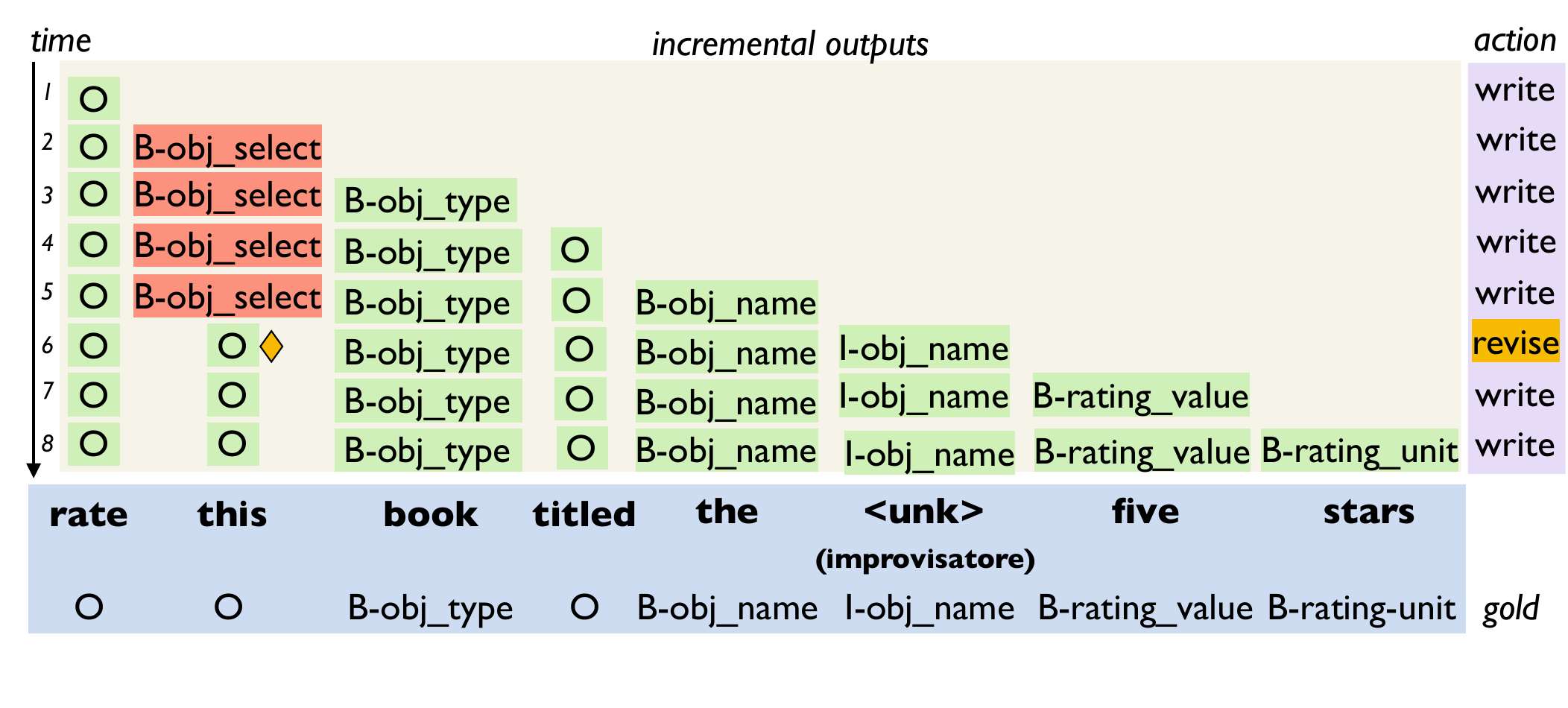}
    \end{subfigure}
	\caption{Additional inference examples from SF-SNIPS obtained with \textsc{Tapir}-Trf. Edited labels are marked by a diamond symbol, with the immediate past output at the top right corner for right-frontier edits. Red labels are incorrect with respect to the final output. In the first example, \textit{how does it} is interpreted as an object name at $t=\{4,5\}$, but is revised to a part of an album when \textsc{Tapir} reads \textit{by}. It still makes a mistake at the last step, as it edits the label for \textit{how} from \texttt{B-album} to \texttt{B-track} when it is unnecessary. \textsc{Tapir} initially labels \textit{this} in \textit{rate this} as \texttt{B-object\_select} in the second example, which probably suits the available evidence at $t=2$. When it encounters the \texttt{UNK} token, \texttt{B-object\_select} is revised to \texttt{O}.}
    \vspace*{-0.5cm}
	\label{fig:appendixqualitative}
\end{figure*}

\begin{figure*}[ht]
\centering

\begin{subfigure}[t]{\textwidth}
   \centering
   \includegraphics[width=0.3\linewidth]{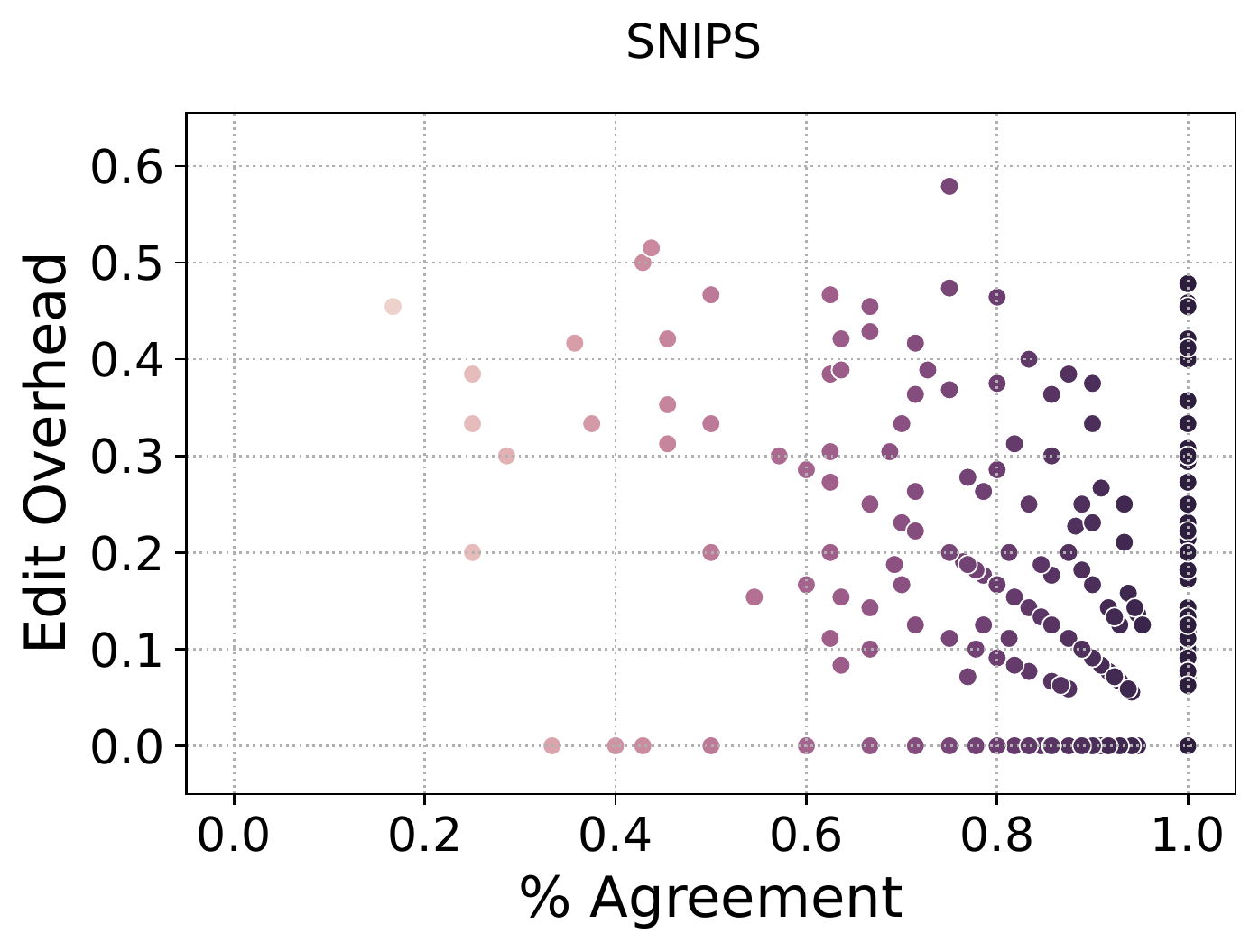}
    \hfill
  \includegraphics[width=0.3\linewidth]{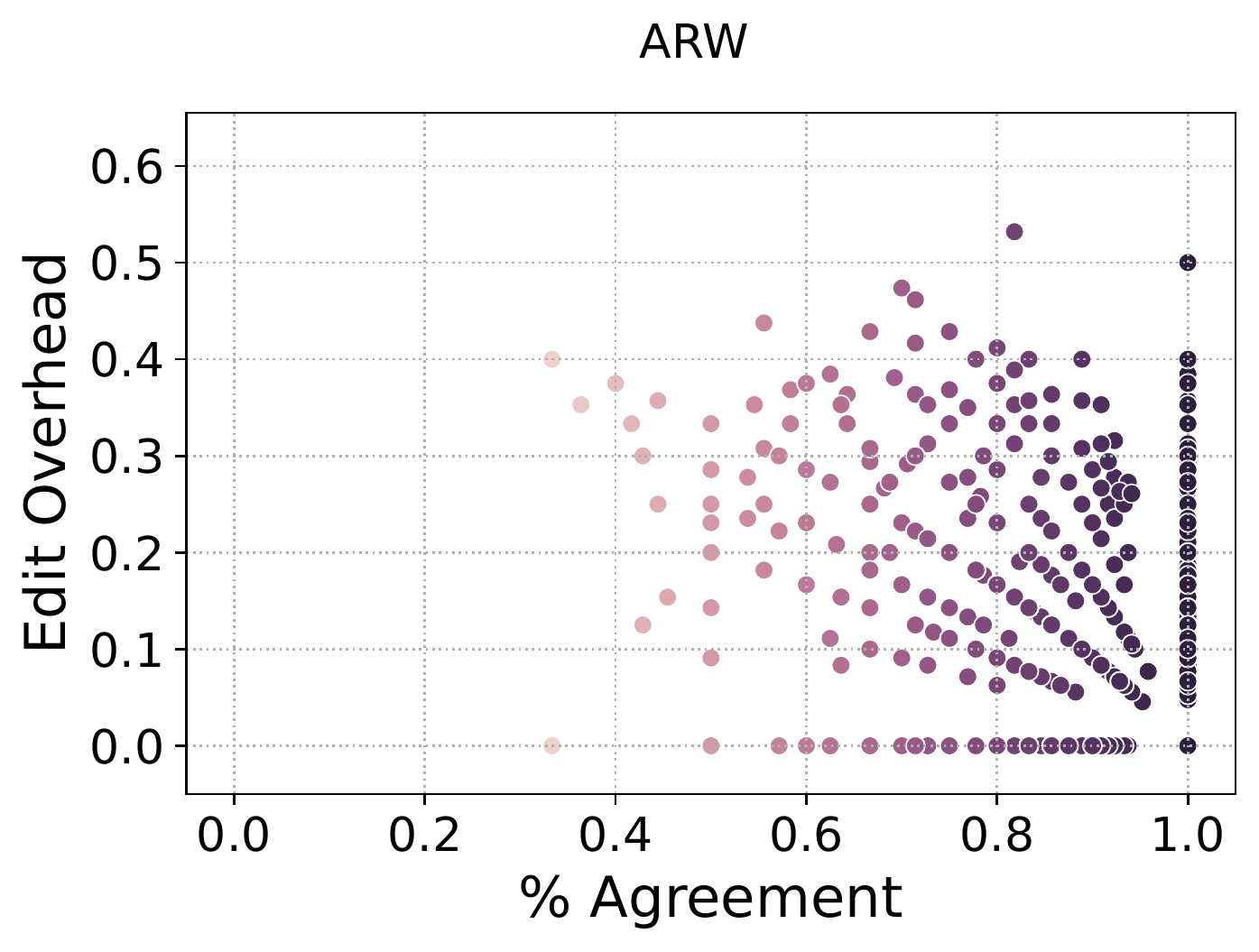}
   \hfill
   \includegraphics[width=0.3\linewidth]{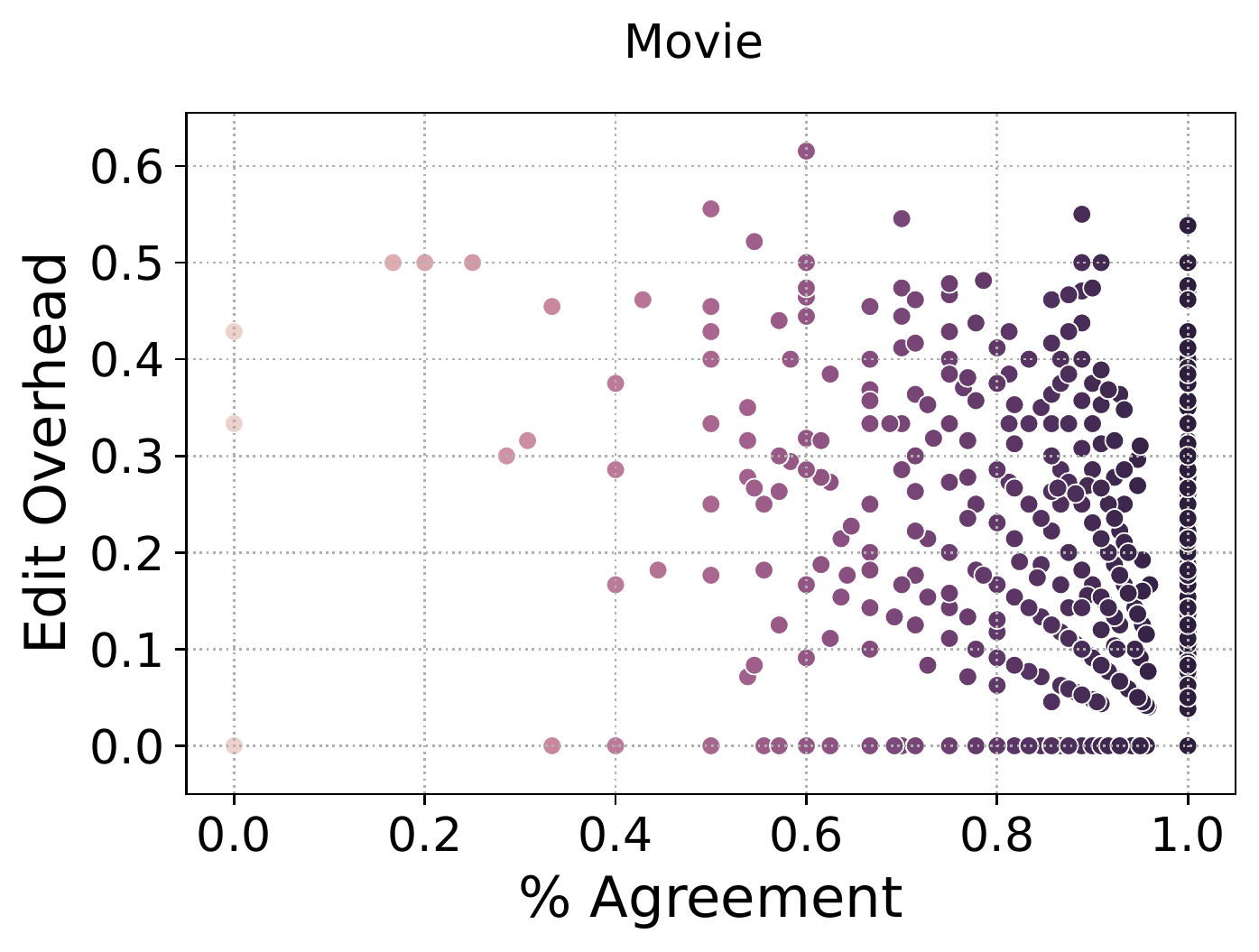}
\end{subfigure}

\begin{subfigure}[t]{\textwidth}
   \centering
   \includegraphics[width=0.3\linewidth]{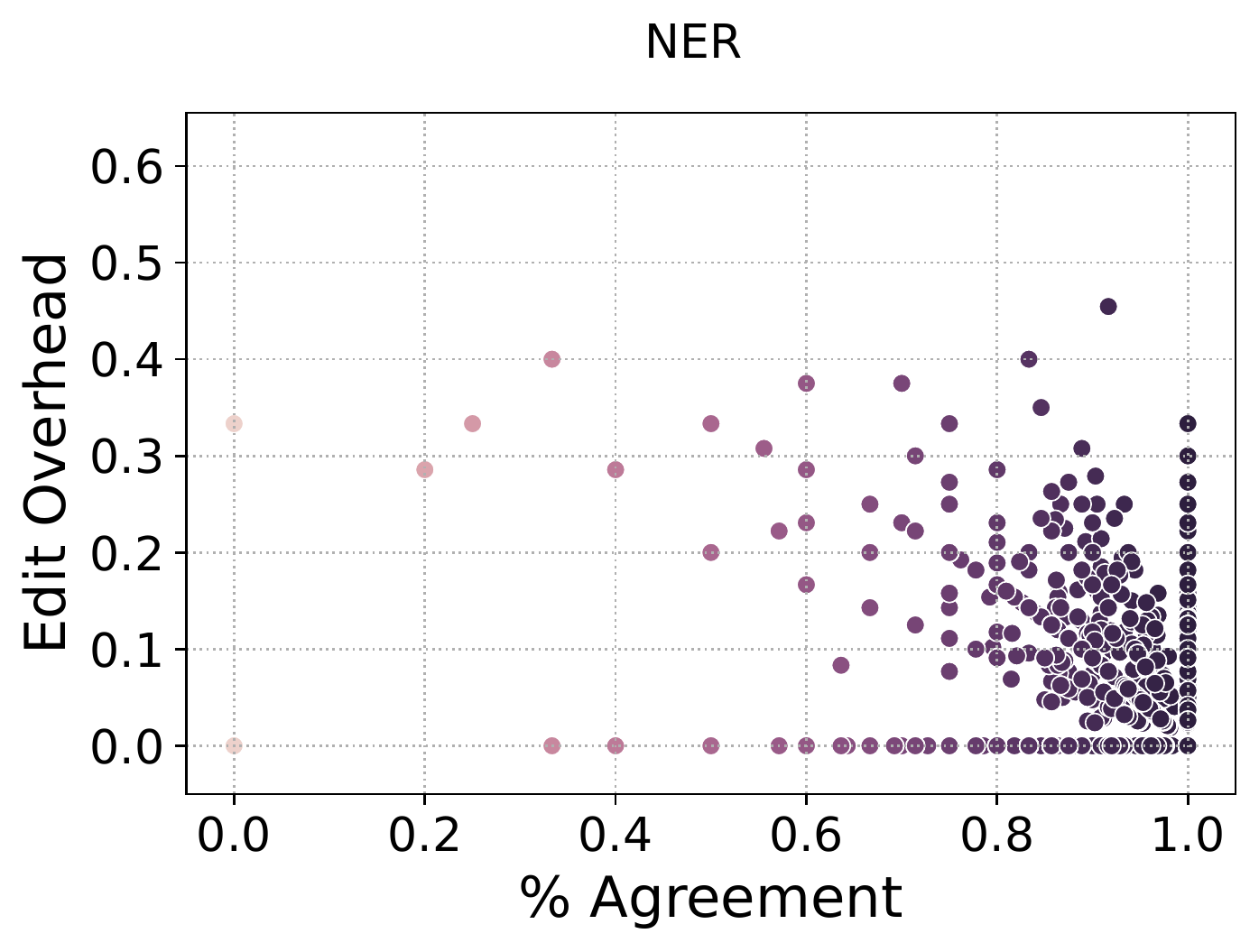}
    \hfill
  \includegraphics[width=0.3\linewidth]{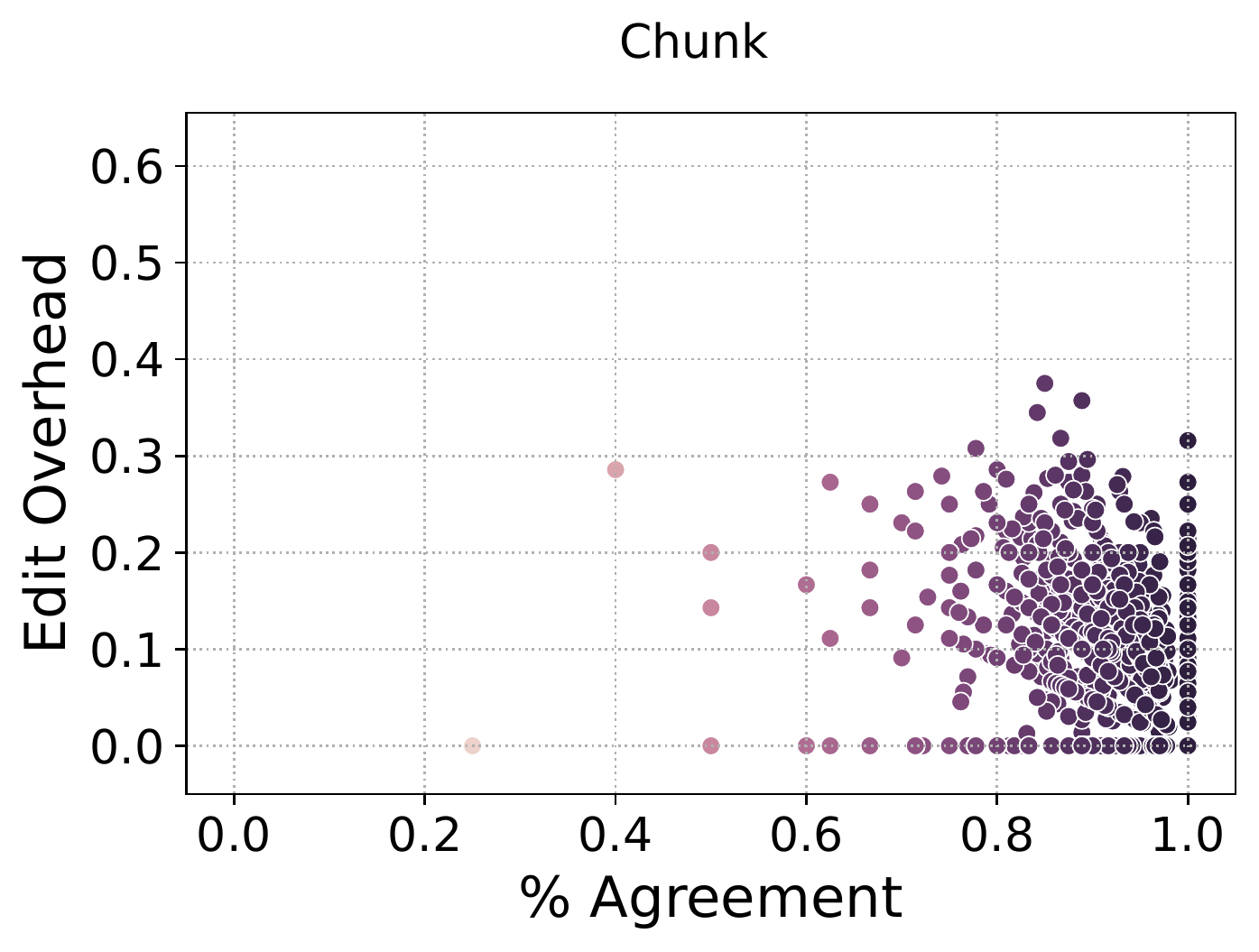}
   \hfill
   \includegraphics[width=0.3\linewidth]{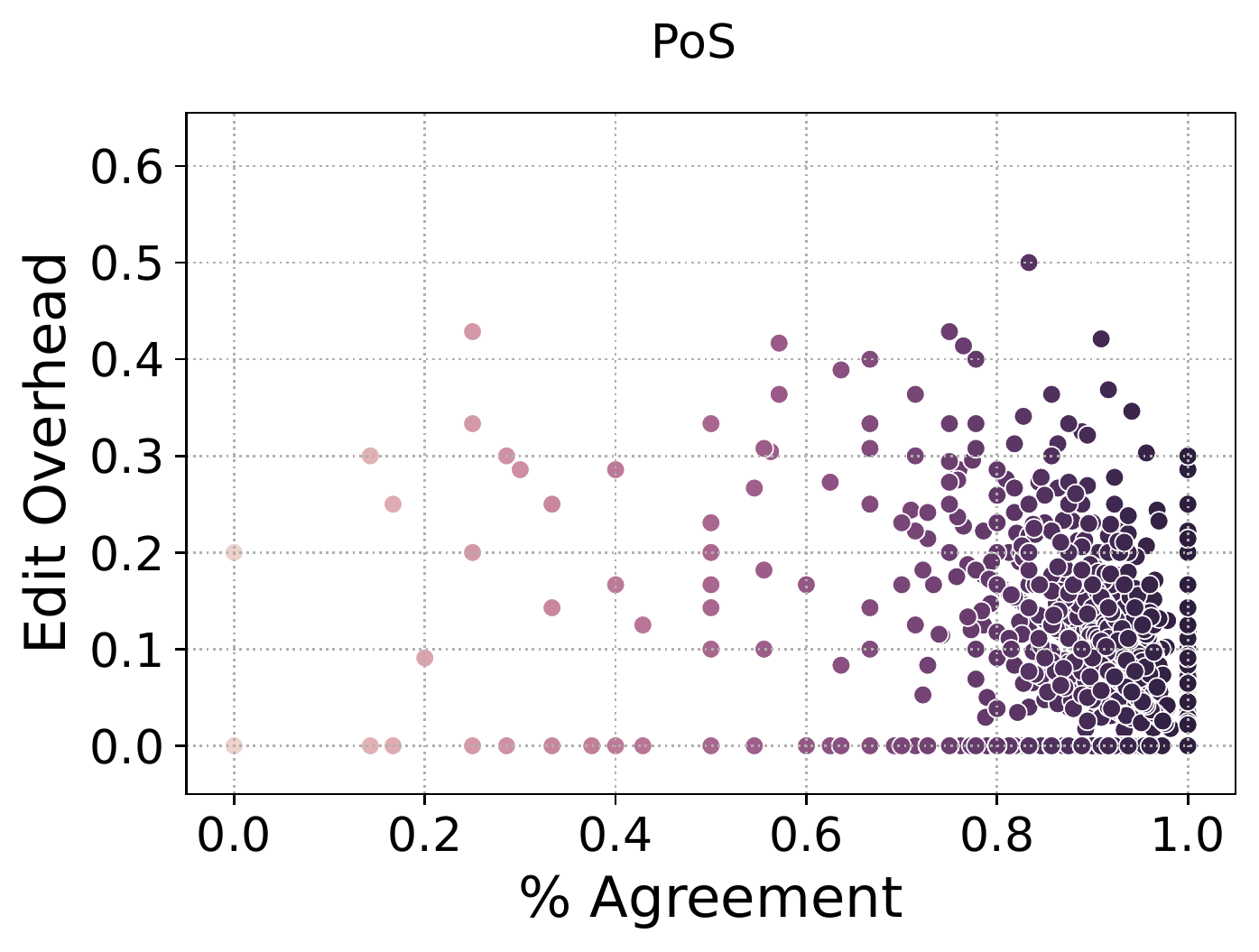}
\end{subfigure}

\begin{subfigure}[t]{\textwidth}
   \centering
   \includegraphics[width=0.3\linewidth]{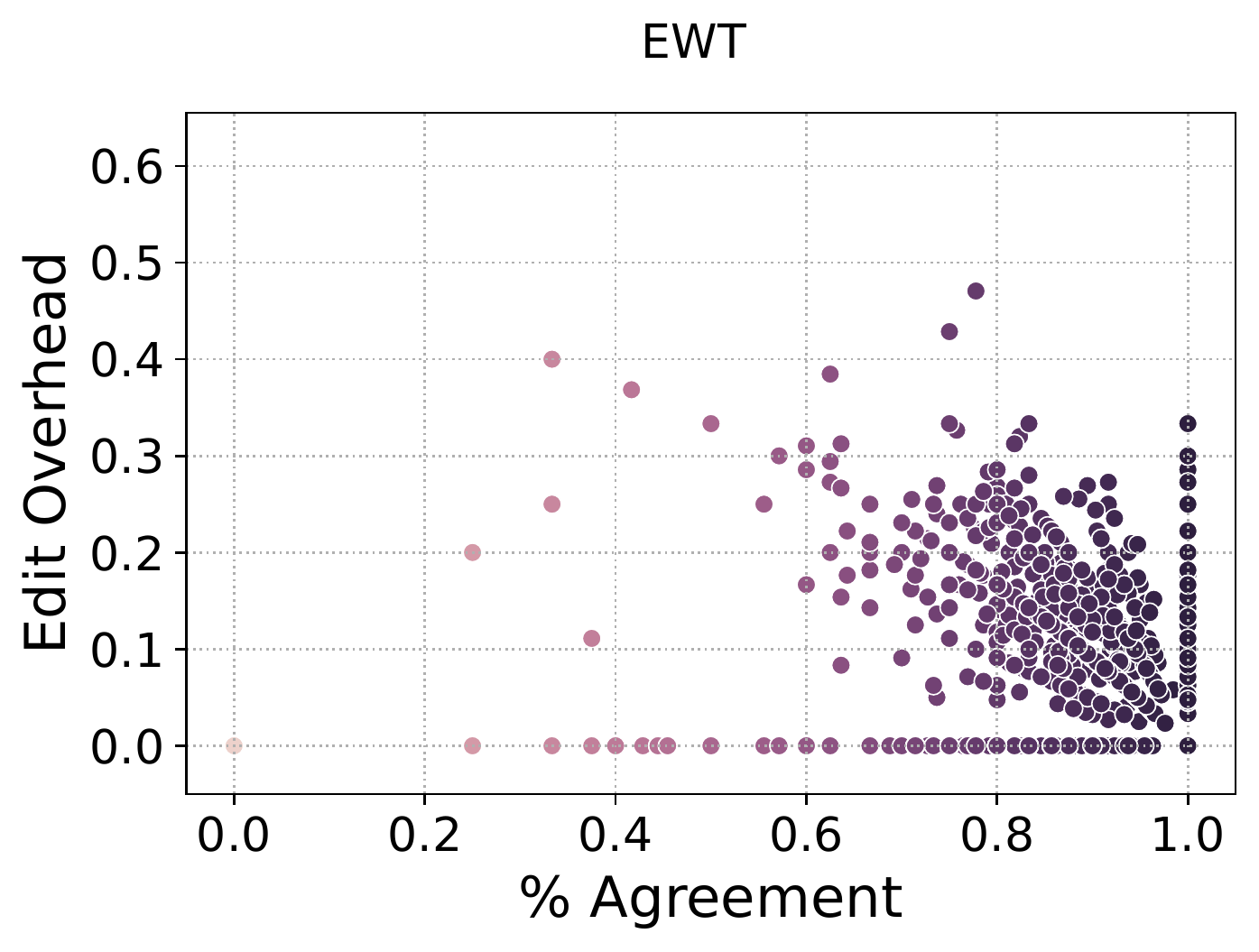}
\end{subfigure}

  \caption{Agreement percentage of the final output between the incremental processor and the reviser of \textsc{Tapir}-Trf. Disagreements are relatively rare and when they disagree, the range of edit overhead is hardly different compared to the case where both components fully agree with each other.}
  \label{fig:agreement} 
\end{figure*}

In this section, we provide information regarding the hyperparameters, implementation, and additional details that are needed to reproduce this work (Table \ref{table:hyperparamsearch} - \ref{table:rawnumbers_threshold}). We also present supplementary materials to accompany the main text (Proof for \S\ref{sec:rev-signal}, Algorithm \ref{alg:twopass}, Figure \ref{fig:appendixqualitative} - \ref{fig:agreement}). 

For all of our experiments, the seed is set to 42119392. We re-implement the Transformer and the LSTMN used in this work, while for the Linear Transformer (LT), we use the official implementation.\footnote{\url{https://linear-transformers.com/}} Further information regarding dependencies and versions are available in the repository.

\subsection*{Datasets}

Tables \ref{table:appendixdatasets1} and \ref{table:appendixdatasets2} summarise the datasets. For SNIPS, we use the preprocessed data and splits provided by \citet{e-etal-2019-novel}. As the MIT Movie dataset does not have an official validation set, we randomly select 10\% of the training data as the validation set. We also remove sentences longer than 200 words. While we use the validation set to tune the hyperparameters of our models, the results on test sets are obtained by using models that are trained on the combination of training and validation sets. 

\subsection*{Action Sequence Generation}

For the action sequence generation, we train a single-layer LT for 20 epochs with linear learning rate warm-up over the first 5 epochs. We use AdamW optimiser \citep{loshchilov2018decoupled} with $\beta_1 = 0.9$ and $\beta_2 = 0.98$. Xavier initialisation \citep{Glorot10understandingthe} is applied to all parameters. The learning rate is set to $1e^{-4}$, with gradient clipping of 1, dropout of 0.1, and batch size of 128. We set the FFNN dimension to 2048 and self-attention dimension to 512, with 8 attention heads. The same hyperparameters are used for all datasets. Action sequences for training the final models are obtained using single-layer LTs that are trained on the combination of training and validation sets.

\subsection*{Implementation and training details}
Our reference model and \textsc{Tapir} are trained for 50 epochs with dropout of 0.1 and early stopping with patience of 10. For AdamW, we use $\beta_{1} = 0.9$ and $\beta_{2} = 0.98$. We also apply Xavier initialisation to all parameters. To train the reference model and the reviser, we use linear learning rate warmup over the first 5 epochs. The learning rate is decayed by 0.5 after 30, 40, and 45 epochs for all models. The number of attention heads for Transformer and LT encoders is set to 8, where each head has the dimension of $d_{\text{model}}/8$ and $d_{\text{model}}$ is the self-attention dimension. The embedding projection layer is of size $d_{\text{model}}$. For OOV words, we follow \citet{lectrack} by randomly replacing tokens with an \texttt{UNK} token during training with a probability that we set to 0.02, and then use this token whenever we encounter unknown words during inference. Hyperparameter search is performed using Optuna \citep{optuna_2019} by maximising the corresponding non-incremental metric on the validation set. We limit the hyperparameter search trials to 25 for all of our experiments. Different from the two-pass model in \citet{sainath2019twopass}, during training we do not take the trained reviser in step (\ref{train1}), freeze its weights, and use it for training step (\ref{train2}). This is because when recomputation occurs, we use output logits from the reviser to recompute $z$ and $\varphi$, but this would mean that the error from the previous $z$ and $\varphi$ cannot be backpropagated. We also experimented using unit logits $\left( \tilde{y}/\|\tilde{y}\|\right)$ to compute $z$, as the logits value from the incremental processor and the reviser might differ in magnitude, but using raw logits proved to be more effective. All the experiments were conducted on a GeForce GTX 1080 Ti and took $\sim$2 weeks to complete.

\begin{table}[htbp]{
\footnotesize
\captionsetup{singlelinecheck = false, justification=justified}
\begin{subtable}[h]{\columnwidth}
\centering
\begin{tabular}{l l} 
\toprule
\multicolumn{1}{l}{Hyperparameters}  & \\
\midrule
Layers & 1, 2, 3, 4 \\
Gradient clipping & no clip, 0.5, 1\\
Learning rate & $5e^{-5}, 7e^{-5}, 1e^{-4}$\\
Batch size & 16, 32, 64\\
Feed-forward dimension & 1024, 2048 \\
Self-attention dimension & 256, 512 \\
\bottomrule
\end{tabular}
\captionsetup{singlelinecheck = false, justification=centering}
\caption{Search space for the reference model}
\vspace*{.1cm}
\label{tab:hyperparamsearch1}
\end{subtable}

\begin{subtable}[h]{\columnwidth}
\begin{tabular}{l l} 
\toprule
\multicolumn{1}{l}{Hyperparameters}  & \\
\midrule
\bigcell{l}{Incremental Processor \\ \& Controller} \\
\cmidrule{1-1}
LSTM layers & 1, 2, 3, 4\\
Controller layers & 1, 2\\
Gradient clipping & no clip, 0.5, 1 \\
Learning rate & $5e^{-5}, 7e^{-5}, 1e^{-4}, 1e^{-3}$\\ 
Batch size & 16, 32, 64\\
LSTM hidden dimension & 256, 512\\
Controller hidden dimension & 256, 512\\
Memory size & 3, 5, 7 \\
\midrule
Reviser \\
\cmidrule{1-1}
Layers & 1, 2, 3, 4 \\
Gradient clipping & no clip, 0.5, 1\\
Learning rate & $5e^{-5}, 7e^{-5}, 1e^{-4}$\\
Batch size & 16, 32, 64\\
Feed-forward dimension & 1024, 2048 \\
Self-attention dimension & 256, 512 \\
\bottomrule
\end{tabular}
\captionsetup{singlelinecheck = false, justification=centering}
\caption{Search space for \textsc{Tapir}}
\label{tab:hyperparamsearch2}
\end{subtable}

\caption{Hyperparameter search space for the reference model and \textsc{Tapir}. The reference model and the reviser share the same search space.}
\label{table:hyperparamsearch}
}
\end{table}

\subsection*{Overview of the Linear Transformer}

The Linear Transformer (LT) \citep{pmlr-v119-katharopoulos20a} uses kernel-based formulation and associative property of matrix products to approximate the softmax attention in conventional Transformers, which is a special case of self-attention. In LT, the self-attention for the $i$-th position is expressed as: 

\vspace*{-0.7cm}

\begin{align}
    \text{Att}_{i}(Q, K, V) &= \frac{\phi (Q_{i})^{\top}S_{p}}{\phi (Q_{i})^{\top}Z_{p}} \\
    S_{p} = \sum_{j=1}^{p}\phi (K_{j})V_{j}^{\top}&; Z_{p} = \sum_{j=1}^{p}\phi (K_{j}) \label{eq:state} 
\end{align}

\vspace*{-0.2cm}

For unmasked attention with a sequence length of $N$, $p=N$ whereas $p=i$ for causal attention. The feature map $\phi$ is an exponential linear unit (elu) \citep{DBLP:journals/corr/ClevertUH15}, specifically  $\phi(x) = \text{elu}(x) + 1$. LT can be viewed as an RNN with hidden states $S$ and $Z$ that are updated as follows:

\vspace*{-0.6cm}

\begin{align}
    S_{i} &= S_{i-1} + \phi(K_{i})V_{i}^{\top} \label{state1} \\
    Z_{i} &= Z_{i-1} + \phi(K_{i}) \label{state2}
\end{align}

\vspace*{-0.2cm}

with initial states $S_{0} = Z_{0} = 0$.

\subsection*{Proof: Duality of the Linear Transformer}
Ideally, the information regarding when to revise should be obtained with RNNs, as they have properties that are crucial for incremental processing and therefore can capture high-quality supervision signal. In practice, this is difficult because it cannot perform revision and its recurrence only allows a unidirectional information flow, which prevents a backward connection to any past outputs. For example, creating a link between the input $x_{t}$ and any past outputs requires computing past hidden states from $h_{t}$, which is non-trivial. One technique to achieve this is to use reversible RNNs \citep{reversiblernn} to reverse the hidden state transition, but this is only possible during training. Another 
approach involves using neural ODE \citep{neuralode} to solve the initial value problem from $h_{0}$, which yields $h_{t}$ for any time step $t$ as the solution, but it would be just an approximation of the true hidden state. 

Let us consider an RNN in an incremental scenario, keeping a hidden state $h_{j}$. How does $x_{t}$ affect the \textit{earlier} output $y_{j}$ for $ 1 \leq j < t$? We want an answer that satisfies the following conditions for incremental processing: 

\begin{enumerate}
   \item \label{cond1} The converse hidden state for time step $j$ computed at time step $t$, $\ddot{h}_{j}$, is a function of $x_{t}$.
   \item \label{cond2} The computation of $h_{t}$ is a function of $h_{t-1}$, and not of $\ddot{h}_{t-1}$. This is consistent with how RNNs work.
   \item \label{cond3} The computation of $h_{t-1}$ is valid iff it involves hidden states $h_{0}, \dots, h_{t-2}$ that agree with condition (\ref{cond2}) in their corresponding step.
\end{enumerate}

In other words, we want a way to compute converse states $\ddot{h}_{j}$ as a function of $x_{t}$, but it should not be affecting $h_{t}$, which is only supposed to be computed using past hidden states built from left to right. We are able to satisfy the conditions above and resolve the conflicting hidden state computation by using the Linear Transformer (LT) \citep{pmlr-v119-katharopoulos20a}, which can be viewed both as a Transformer and as an RNN. This allows us to get the supervision signal to determine when revision should happen through restart-incremental computation, while still observes how $x_{t}$ affects all past outputs from the perspective of RNNs.

Let us consider the self-attention computation at time step $t$ for the current and past positions $n, n-1, n-2; n=t$ obtained with a LT under restart-incrementality:

\vspace*{-0.5cm}

\begin{align}
    \text{Att}_{n}^{t}(Q, K, V) &= \frac{\phi (Q_{n})^{\top}S_{n}}{\phi (Q_{n})^{\top}Z_{n}} \label{eq:cond2_after} \\
    \text{Att}_{n-1}^{t}(Q, K, V) &= \frac{\phi (Q_{n-1})^{\top}S_{n}}{\phi (Q_{n-1})^{\top}Z_{n}} \label{eq:cond1_t_1}\\
    \text{Att}_{n-2}^{t}(Q, K, V) &= \frac{\phi (Q_{n-2})^{\top}S_{n}}{\phi (Q_{n-2})^{\top}Z_{n}} \label{eq:cond1_t_2}
\end{align}

\vspace*{-0.1cm}

From equations (\ref{eq:cond1_t_1}) and (\ref{eq:cond1_t_2}) we can see that the hidden state $S$ for computing the representations at positions $n-1$ and $n-2$ are functions of $x_{n}$ which satisfies condition (\ref{cond1}). Furthermore, they are equal to each other \emph{i.e.,} $\ddot{S}_{n-2} = \ddot{S}_{n-1} = S_{n} = S^{t}$. Note that we only consider $S$, however the proof also holds for $Z$. To satisfy condition (\ref{cond2}), consider the self-attention at time step $t-1$ for position $n-1$:

\vspace*{-0.5cm}

\begin{align}
    \text{Att}^{t-1}_{n-1}(Q, K, V) &= \frac{\phi (Q_{n-1})^{\top}S_{n-1}}{\phi (Q_{n-1})^{\top}Z_{n-1}} \label{eq:cond2_before}
\end{align}

\vspace*{-0.1cm}

$S^{t}$ is a function of $S^{t-1}$ in equation (\ref{eq:cond2_before}), $S^{t} = S^{t-1} + \phi(K_{n})V_{n}^{\top}$. We also know that $S^{t} = \ddot{S}_{n-1}$, which means that condition (\ref{cond2}) is not completely fulfilled. However, the last clause can be relaxed as it only exists to ensure that the incremental assumption during the computation of $S^{t}$ is met. The reason for this is because there are two ways to view the computation of $S$ at any time step $t$: (1) by updating the previous state $S^{t-1}$, or (2) computing $S^{t}$ directly from input tokens $x_{1},\dots, x_{n=t}$, which is analogous to the kernel trick, but in this case $S$ is a combination of projected input tokens. The latter view can be used to relax condition (\ref{cond2}), as it means $S^{t}$ does not completely depend on the previous state ($S^{t-1}$ or $\ddot{S}_{n-1}$) like in conventional RNNs, but can also be computed directly from input tokens while still obeying incremental assumptions.

Fulfilling condition (\ref{cond3}) requires that condition (\ref{cond2}) holds for all preceding time steps. Formally, $S^{i} = f(S^{i-1})$ and $S^{i} \neq f(\ddot{S}); 1 \leq i \leq t-1$. This is satisfied by the fact that $S^{i} = S^{i-1} + \phi(K_{i})V_{i}^{\top}$ and taking the perspective of $S$ as a combination of projected input tokens for relaxation. Notice that equation (\ref{eq:cond2_after}) is causal and can be expressed as an RNN at time step $t$ while equations (\ref{eq:cond1_t_1}) and (\ref{eq:cond1_t_2}) are acausal. This proof only holds for a \emph{single} layer of LT due to how information flows between layers. Let us consider the computation of $S$ for a multi-layer LT. At time step $t$, we compute $S_{n, l}^{t}$ for position $n=t$ in layer $l$ using $x_{1}^{l}, \dots, x_{n}^{l}$, which are outputs of layer $l-1$. At the same time, these inputs for layer $l$ are computed using $S^{t}_{n, l-1}$ from layer $l-1$. This means $x_{1}^{l}, \dots, x^{l}_{n-1}$ are functions of $x^{l-1}_{n}$, which violates the incremental assumption for the input. Therefore, we will be unable to properly examine the effect of the current input on all past outputs if we employ a multi-layer LT.

\begin{algorithm*}
\caption{\textsc{Tapir}}\label{alg:twopass}
\begin{algorithmic}[1]
\Require Incremental processor $\psi$, reviser $\eta$, caches $\Gamma^{h}, \Gamma^{z}, \Gamma^{p}$, controller $\xi$, policy $\pi_{\theta}$, input $X$, input buffer $X_{buf}$, output buffer $Y_{buf}$
\State \textbf{Initialise:} $h_{0} \gets 0$, $x_{1} \Leftarrow X$, $\tilde{k}_{1} \gets 0$, $\tilde{c}_{1} \gets 0$, $t \gets 1$

\While{$t \leq |X|$}
\State $h_{t} \gets \psi(h_{t-1}, x_{t})$, $\tilde{y}_{t} \gets f_{\psi}(h_{t})$, $y_{t} \gets \text{softmax}(\tilde{y}_{t})$
\If{$\Gamma^{p} \neq \emptyset$}
\For{$i \gets 1$ to $\min{(t-1, N)}$}
\State $\gamma_{i}^{p} \Leftarrow \Gamma^{p}$, $e_{i}^{t} \gets f_{\xi}(\gamma_{i}^{p}, h_{t}, \tilde{k}_{t-1})$
\EndFor
\State $s^{t} \gets \text{softmax}(e^{t})$, $\tilde{k}_{t} \gets \sum_{i}s_{i}^{t}\gamma_{i}^{p}$, $\tilde{c}_{t} \gets \sum_{i}s_{i}^{t}c_{i + \max{(0, t-N-1)}}$
\EndIf
\State $k_{t}, c_{t} \gets \xi(\tilde{k}_{t}, \tilde{c}_{t}, x_{t})$
\State $a_{t} \gets \pi_{\theta}(k_{t})$, $X_{buf} \Leftarrow x_{t}$
\If{$|\Gamma^{h}| = N$} 
\State \textbf{del} $\gamma_{1}^{h}$ \Comment{Discard the first cache element when full.}
\EndIf
\State $\Gamma^{h} \Leftarrow h_{t}$ \Comment{Update the cache.}
\If{$a_{t}$ = \texttt{WRITE}}
\State $Y_{buf} \Leftarrow y_{t}$, $z \gets f_{z}(\tilde{y}_{t})$, $\varphi \gets f_{\phi}(h_{t}, z)$
\If{$|\Gamma^{z}| = N$ \textbf{and} $|\Gamma^{p}| = N$}
\State \textbf{del} $\gamma_{1}^{z}$, \textbf{del} $\gamma_{1}^{p}$
\EndIf
\State $\Gamma^{z} \Leftarrow z$, $\Gamma^{p} \Leftarrow \varphi$
\ElsIf{$a_{t}$ = \texttt{REVISE}}
\State $\tilde{y}_{\leq t}^{\eta} \gets f_{\eta}(\eta({X_{buf}}))$, $Y_{buf} \gets \text{softmax}(\tilde{y}_{\leq t}^{\eta})$, $\Gamma^{z} \gets \emptyset$, $\Gamma^{p} \gets \emptyset$
\For{$j \gets \max{(1, t-N + 1)}$ to $t$}
\State $h_{j} \Leftarrow \Gamma^{h}$, $z \gets f_{z}(\tilde{y}_{j}^{\eta})$, $\varphi \gets f_{\phi}(h_{j}, z)$
\State $\Gamma^{z} \Leftarrow z$, $\Gamma^{p} \Leftarrow \varphi$
\EndFor
\EndIf
\State $x_{t+1} \Leftarrow X$, $t \gets t + 1$
\EndWhile
\end{algorithmic}
\end{algorithm*}

\begin{table*}[ht]{
\footnotesize
\setlength\tabcolsep{4.5pt}
\begin{subtable}[h]{\textwidth}
\centering
\begin{tabular}{l c c c c c c} 
\toprule
\multicolumn{1}{l}{Tasks/Models}  & \multicolumn{1}{l}{Layers} & \multicolumn{1}{c}{Clip} & \multicolumn{1}{l}{Learning Rate} & \multicolumn{1}{l}{Batch} & \multicolumn{1}{l}{ Feed-forward} & \multicolumn{1}{l}{Self-attention}\\
\midrule
\bigcell{l}{Reference model \& \\ Transformer reviser} \\
\cmidrule{1-1}
SF-SNIPS & 4 & no clip & $1e^{-4}$ & 16 & 2048 & 512\\
SF-ARW & 4 & 0.5 & $1e^{-4}$ & 16 & 1024 & 256\\
SF-Movie & 4 & no clip & $5e^{-5}$ & 16 & 2048 & 256\\
NER-CoNLL & 3 & 1 & $1e^{-4}$ & 64 & 2048 & 512\\
Chunk-CoNLL & 3 & 0.5 & $7e^{-5}$ & 32 & 2048 & 512\\
PoS-CoNLL & 3 & no clip & $1e^{-4}$ & 16 & 2048 & 512\\
PoS-UD-EWT & 2 & -1 & $7e^{-5}$ & 16 & 2048 & 512\\
\midrule
LT reviser \\
\cmidrule{1-1}
SF-SNIPS & 3 & 0.5 & $1e^{-4}$ & 32 & 1024 & 512\\
SF-ARW & 4 & 1 & $1e^{-4}$ & 32 & 2048 & 512\\
SF-Movie & 4 & 0.5 & $1e^{-4}$ & 16 & 1024 & 512\\
NER-CoNLL & 3 & 0.5 &  $1e^{-4}$ & 16 & 2048 & 512\\
Chunk-CoNLL & 4 & 0.5 & $1e^{-4}$ & 16 & 1024 & 512\\
PoS-CoNLL & 1 & no clip & $1e^{-4}$ & 16 & 2048 & 512\\
PoS-UD-EWT & 3 & 0.5 & $7e^{-5}$ & 16 & 2048 & 512\\
\bottomrule
\end{tabular}
\caption{Reference model, Transformer and LT revisers}
\end{subtable}

\vspace*{.1cm}

\setlength\tabcolsep{3.5pt}
\begin{subtable}[h]{\textwidth}
\centering
\begin{tabular}{l c c c c c c c c} 
\toprule
  & \multicolumn{2}{c}{Layers} & & & & \multicolumn{2}{c}{Dimension} & \\ \cmidrule(lr){2-3} \cmidrule(lr){7-8}
\multicolumn{1}{l}{Tasks/Models}  & \multicolumn{1}{l}{LSTM} & \multicolumn{1}{c}{Ctrl.} & \multicolumn{1}{c}{Clip} & \multicolumn{1}{l}{Learning Rate} & \multicolumn{1}{l}{Batch} & \multicolumn{1}{l}{LSTM} & \multicolumn{1}{c}{Ctrl.} & \multicolumn{1}{l}{Memory}\\
\midrule
Transformer reviser \\
\cmidrule{1-1}
SF-SNIPS & 1 & 1 & no clip & $1e^{-3}$ & 16 & 512 & 256 & 5\\
SF-ARW & 1 & 2 & no clip & $1e^{-4}$ & 32 & 512 & 256 & 7\\
SF-Movie & 1 & 1 & no clip & $7e^{-5}$ & 16 & 512 & 512 & 7\\
NER-CoNLL & 3 & 1 &  0.5 & $1e^{-3}$ & 16 & 256 & 512 & 3\\
Chunk-CoNLL & 1 & 2 & 1 & $1e^{-3}$ & 32 & 512 & 256 & 5\\
PoS-CoNLL & 1 & 1 & no clip & $5e^{-5}$ & 16 & 256 & 256 & 7\\
PoS-UD-EWT & 1 & 2 & 0.5 & $1e^{-3}$ & 16 & 512 & 256 & 3\\
\midrule
LT reviser \\
\cmidrule{1-1}
SF-SNIPS & 2 & 2 & no clip & $1e^{-3}$ & 64 & 256 & 512 & 3\\
SF-ARW & 1 & 1 & 1 & $7e^{-5}$ & 64 & 256 & 256 & 5\\
SF-Movie & 1 & 1 & 1 & $1e^{-4}$ & 16 & 512 & 256 & 7\\
NER-CoNLL & 1 & 1 &  no clip & $1e^{-3}$ & 16 & 256 & 512 & 5\\
Chunk-CoNLL & 2 & 1 & 1 & $1e^{-3}$ & 16 & 512 & 512 & 7\\
PoS-CoNLL & 1 & 2 & 1 & $5e^{-5}$ & 16 & 256 & 512 & 3\\
PoS-UD-EWT & 1 & 1 & no clip & $1e^{-3}$ & 16 & 256 & 256 & 3\\
\bottomrule
\end{tabular}
\caption{Incremental processor and controller}
\end{subtable}
\captionsetup{singlelinecheck = false, justification=justified}
\caption{Hyperparameters for our experiments. We use the same hyperparameters for the delayed variants.}
\label{table:baselinehyperparam}
}
\end{table*}

\begin{table*}[ht]{
\footnotesize
\setlength\tabcolsep{5pt}
\captionsetup{singlelinecheck = false, justification=justified}
\begin{center}
\begin{tabular}{l l p{3cm} p{3.5cm} p{2cm}} 
\toprule
Tasks & Dataset & Publication & License & Downloadable \\
\midrule
Slot filling & SNIPS & \citet{coucke2018snips} & 
CC0 & \href{https://github.com/sonos/nlu-benchmark}{link} / \href{https://github.com/ZephyrChenzf/SF-ID-Network-For-NLU/tree/master/data/snips}{preproc.}\\
Slot filling & Alarm, reminder, \& weather & \citet{schuster-etal-2019-cross-lingual} & CC BY-SA & \href{https://fb.me/
multilingual_task_oriented_data}{link} \\
Slot filling & MIT Movie, eng corpus & \citet{mitrestaurantmovie} & - & \href{https://groups.csail.mit.edu/sls/downloads/}{link} \\
NER & \multirow{3}{*}{CoNLL-2003} & \multirow{3}{3cm}{\citet{tjong-kim-sang-de-meulder-2003-introduction}}  & \multirow{3}{3.5cm}{text: NIST research agreement; annotation: -} & \multirow{3}{2cm}{\href{https://www.clips.uantwerpen.be/conll2003/ner/}{link}} \\
Chunking &  &  & & \\
PoS tagging &  &  & & \\
PoS tagging & Universal Dependencies, EWT & \citet{silveira-etal-2014-gold}  & CC BY-SA 4.0 
& \href{https://github.com/UniversalDependencies/UD_English-EWT}{link}\\

\bottomrule
\end{tabular}
\end{center}
\vspace*{-0.2cm}
\caption{Details about each dataset.}
\label{table:appendixdatasets1}
}
\end{table*}

\begin{table*}[ht]{
\footnotesize
\setlength\tabcolsep{5pt}
\captionsetup{singlelinecheck = false, justification=justified}
\begin{center}
\begin{tabular}{l r r r r r c r r c r } 
\toprule
& \multicolumn{3}{c}{No. of Seq.} & & & \multicolumn{2}{c}{Token Size} & & \multicolumn{2}{c}{Avg. Seq. Length}\\
\cmidrule{2-4} \cmidrule{7-8} \cmidrule{10-11}
\multicolumn{1}{l}{Tasks} & \multicolumn{1}{l}{Train} & \multicolumn{1}{l}{Valid} & \multicolumn{1}{c}{Test} & \multicolumn{1}{l}{Labels} & \multicolumn{1}{l}{Vocab Size} &  \multicolumn{1}{l}{Train \& Valid} &  \multicolumn{1}{c}{Test} & &  \multicolumn{1}{l}{Train \& Valid} &  \multicolumn{1}{c}{Test}\\
\midrule
SF-SNIPS & 13,084 & 700 & 700 & 72 & 11,765 & 124,084 & 6,354 & & \hphantom{0}9.002 & 9.077\\
SF-ARW & 30,521 & 4,181 & 8,621 & 28 & 4,215 & 251,915 & 62,591 & & \hphantom{0}7.259 & 7.260\\
SF-Movie & 8,797 & 978 & 2,443 & 25 & 6,710 & \hphantom{0}99,491 & 24,686 & & 10.178 & 10.105\\
NER-CoNLL & 14,041 & 3,249 & 3,452 & 9 & 26,882 & 254,979 & 46,394 & & 14.747 & 13.440\\
Chunk-CoNLL & 14,041 & 3,249 & 3,452 & 23 & 26,882 & 254,979 & 46,394 & & 14.747 & 13.440\\
PoS-CoNLL & 14,041 & 3,249 & 3,452 & 47 & 26,882 & 254,979 & 46,394 & & 14.747 & 13.440\\
PoS-UD-EWT & 12,543 & 2,001 & 2,077 & 18 & 21,917 & 232,741 & 25,456 & & 16.003 & 12.256\\

\bottomrule
\end{tabular}
\end{center}
\vspace*{-0.2cm}
\caption{Descriptive statistics of the datasets. The vocabulary size is computed from training and validation sets.}
\label{table:appendixdatasets2}
}
\end{table*}

\begin{table*}[ht]{
\footnotesize
\setlength\tabcolsep{18pt}
\begin{subtable}[h]{\columnwidth}
\centering
\begin{tabular}{l c c} 
\toprule
\multicolumn{1}{l}{Tasks}  & \multicolumn{1}{l}{\texttt{WRITE}} & \multicolumn{1}{l}{\texttt{REVISE}}\\
\midrule
SF-SNIPS & 0.763 & 0.237 \\
SF-ARW & 0.831 & 0.169 \\
SF-Movie & 0.764 & 0.236 \\
NER-CoNLL & 0.904 & 0.096 \\
Chunk-CoNLL & 0.838 & 0.162 \\
PoS-CoNLL & 0.702 & 0.298 \\
PoS-UD-EWT & 0.785 & 0.215 \\
\bottomrule
\end{tabular}
\caption{Training sets}
\label{tab:meanwriterevise1}
\end{subtable}
\hfill
\begin{subtable}[h]{\columnwidth}
\centering
\begin{tabular}{l c c} 
\toprule
\multicolumn{1}{l}{Tasks}  & \multicolumn{1}{l}{\texttt{WRITE}} & \multicolumn{1}{l}{\texttt{REVISE}}\\
\midrule
SF-SNIPS & 0.794 & 0.206 \\
SF-ARW & 0.838 & 0.162 \\
SF-Movie & 0.772 & 0.228 \\
NER-CoNLL & 0.910 & 0.090 \\
Chunk-CoNLL & 0.790 & 0.210 \\
PoS-CoNLL &0.819  & 0.181 \\
PoS-UD-EWT & 0.771 & 0.229 \\
\bottomrule
\end{tabular}
\caption{Training and validation sets}
\label{tab:meanwriterevise2}
\end{subtable}

\caption{Mean of \texttt{WRITE} and \texttt{REVISE} action ratios per sentence for training sets and combination of training and validation sets. Most of the time, the mean of the \texttt{WRITE} action ratio is higher compared to the \texttt{REVISE} action ratio.}
\label{table:writereviseprop}
}
\end{table*}

\begin{table*}[ht]{
\footnotesize
\captionsetup{singlelinecheck = false, justification=justified}
\begin{subtable}[h]{\textwidth}
\centering
\begin{tabular}{l c c c c c} 
\toprule
  & \multicolumn{5}{c}{Percentage (\%) by \texttt{REVISE} Ratio} \\ \cmidrule(lr){2-6}
\multicolumn{1}{l}{Tasks} & \multicolumn{1}{c}{0-0.2} & \multicolumn{1}{c}{0.2-0.4} & \multicolumn{1}{c}{0.4-0.6} & \multicolumn{1}{c}{0.6-0.8} & \multicolumn{1}{c}{0.8-1} \\
\midrule
SF-SNIPS & 48.65 & 30.30 & 17.56 & \hphantom{0}3.41 & 0.08 \\
SF-ARW & 63.54 & 23.30 & 11.58 & \hphantom{0}1.55 & 0.03 \\
SF-Movie & 47.60 & 34.39 & 15.63 & \hphantom{0}2.30 & 0.09 \\
NER-CoNLL & 79.67 & 15.53 & \hphantom{0}4.15 & \hphantom{0}0.64 & 0.01 \\
Chunk-CoNLL & 61.73 & 27.23 & 10.21 & \hphantom{0}0.81 & 0.01 \\
PoS-CoNLL & 39.17 & 24.63 & 25.33 & 10.37 & 0.51 \\
PoS-UD-EWT & 50.58 & 33.24 & 14.08 & \hphantom{0}2.06 & 0.05 \\
\bottomrule
\end{tabular}
\captionsetup{singlelinecheck = false, justification=centering}
\caption{Training sets}
\vspace*{.1cm}
\label{tab:revisecount1}
\end{subtable}
\hfill
\begin{subtable}[h]{\textwidth}
\centering
\begin{tabular}{l c c c c c} 
\toprule
  & \multicolumn{5}{c}{Percentage (\%) by \texttt{REVISE} Ratio} \\ \cmidrule(lr){2-6}
\multicolumn{1}{l}{Tasks} & \multicolumn{1}{c}{0-0.2} & \multicolumn{1}{c}{0.2-0.4} & \multicolumn{1}{c}{0.4-0.6} & \multicolumn{1}{c}{0.6-0.8} & \multicolumn{1}{c}{0.8-1} \\
\midrule
SF-SNIPS & 54.80 & 28.66 & 13.99 & 2.52 & 0.03 \\
SF-ARW & 65.01 & 22.88 & 10.71 & 1.35 & 0.04 \\
SF-Movie & 49.21 & 34.36 & 14.54 & 1.85 & 0.04 \\
NER-CoNLL & 81.26 & 14.75 & \hphantom{0}3.61 & 0.35 & 0.03 \\
Chunk-CoNLL & 53.20 & 25.00 & 18.63 & 3.13 & 0.03 \\
PoS-CoNLL &  59.35 & 27.64 & 11.09 & 1.84 & 0.08 \\
PoS-UD-EWT & 47.61 & 32.82 & 16.81 & 2.68 & 0.08 \\
\bottomrule
\end{tabular}
\captionsetup{singlelinecheck = false, justification=centering}
\caption{Training and validation sets}
\label{tab:revisecount2}
\end{subtable}

\caption{Distribution of examples in each dataset by their \texttt{REVISE} action ratio for training sets and combination of training and validation sets. Most of the examples in the datasets have considerably low \texttt{REVISE} action ratio ($<0.6$).}
\label{table:revisecount}
}
\end{table*}

\begin{table*}[ht] 
{\footnotesize
\captionsetup{singlelinecheck = false, justification=justified}
\setlength\tabcolsep{6pt}
\centering
\begin{tabular}{l c c c c c c c }
\toprule
 &  & \multicolumn{3}{c}{\textsc{Tapir}-Trf} & \multicolumn{3}{c}{\textsc{Tapir}-LT} \\
    \cmidrule(lr){3-5} \cmidrule(lr){6-8}
Tasks & Ref. Model & - & D1 & D2 & - & D1 & D2\\
\midrule
SF-SNIPS     & 91.42 & 88.26 & 91.10 & 90.59 & 88.12 & 87.80 & 88.75 \\
SF-ARW    & 95.55 & 94.94 & 94.96 & 95.17 & 93.90 & 94.63 & 94.60  \\
SF-Movie    & 85.20 & 84.69 & 84.90 & 84.30 & 84.25 & 84.36 & 84.33 \\
NER-CoNLL    & 84.69 & 80.95 & 84.52 & 84.68 & 82.38 & 82.90 & 82.52  \\
Chunk-CoNLL    & 89.02 & 88.19 & 88.86 & 88.69 & 85.76 & 86.92 & 87.19  \\
\midrule
PoS-CoNLL  & 93.07 & 92.73 & 92.87 & 92.81 & 92.48 & 92.23 & 91.98   \\
PoS-UD-EWT  & 91.88 & 90.35 & 91.33 & 91.67 & 89.99 & 90.96 & 90.69  \\

\bottomrule
\end{tabular}
\caption{Non-incremental performance of the models on validation sets (F1 for the first group, accuracy for the second group).}
\label{table:nonincrementalappendix}
}
\end{table*}

\begin{table*}[ht] 
{\footnotesize
\captionsetup{singlelinecheck = false, justification=justified}
\setlength\tabcolsep{6pt}
\centering
\begin{tabular}{l c c c c c c c }
\toprule
&  & \multicolumn{3}{c}{\textsc{Tapir}-Trf} & \multicolumn{3}{c}{\textsc{Tapir}-LT} \\
    \cmidrule(lr){3-5} \cmidrule(lr){6-8}
Tasks & Ref. Model & - & D1 & D2 & - & D1 & D2 \\
\midrule
SF-SNIPS     & 16.2 & 38.7 & 38.7 & 38.7 & 29.7 & 29.7 & 29.7 \\
SF-ARW    & \hphantom{0}4.4 & 13.6 & 13.6 & 13.6 & 30.7 & 30.7 & 30.7  \\
SF-Movie    & \hphantom{0}7.2 & 21.0 & 21.0 & 21.0 & 25.7 & 25.7 & 25.7 \\
NER-CoNLL    & 16.7 & 44.7 & 44.7 & 44.7 & 43.7 & 43.7 & 43.7  \\
Chunk-CoNLL    & 16.7 & 44.1 & 44.1 & 44.1 & 45.2 & 45.2 & 45.2  \\
PoS-CoNLL  & 16.7 & 42.0 & 42.0 & 42.0 & 33.8 & 33.8 & 33.8   \\
PoS-UD-EWT  & 12.5 & 34.7 & 34.7 & 34.7 & 38.9 & 38.9 & 38.9  \\

\bottomrule
\end{tabular}
\caption{Number of parameters for each model, in millions.}
\label{table:number of params}
}
\end{table*}

\begin{table*}[ht]{
\footnotesize
\setlength\tabcolsep{10pt}
\captionsetup{singlelinecheck = false, justification=justified}
\begin{center}
\begin{tabular}{ l c c c c c c c} 
\toprule
\multicolumn{1}{l}{Tasks/Models} & \multicolumn{1}{l}{EO} & \multicolumn{1}{l}{CT} & \multicolumn{1}{l}{RC} & \multicolumn{1}{l}{EO$\Delta1$} & \multicolumn{1}{l}{EO$\Delta2$} & \multicolumn{1}{l}{RC$\Delta1$} & \multicolumn{1}{l}{RC$\Delta2$} \\
\midrule
SF-SNIPS \\
\cmidrule{1-1}
Ref. Model & 0.181 & 0.100 & 0.750 & 0.081 & 0.046 & 0.887 & 0.933 \\
\textsc{Tapir}-Trf & 0.074 & 0.055 & 0.876 & - & - & - & - \\
\textsc{Tapir}-Trf (D1) & - & 0.034 & - & 0.046 & - & 0.919 & - \\
\textsc{Tapir}-Trf (D2) & - & 0.030 & - & - & 0.049 & - & 0.922 \\
\textsc{Tapir}-LT & 0.081 & 0.058 & 0.866 & - & - & - & - \\
\textsc{Tapir}-LT (D1) & - & 0.042 & - & 0.057 & - & 0.899 & - \\
\textsc{Tapir}-LT (D2) & - & 0.036 & - & - & 0.053 & - & 0.905 \\
\midrule
SF-ARW \\
\cmidrule{1-1}
Ref. Model & 0.079 & 0.041 & 0.906 & 0.017 & 0.008 & 0.977 & 0.989 \\
\textsc{Tapir}-Trf & 0.032 & 0.019 & 0.950 & - & - & - & - \\
\textsc{Tapir}-Trf (D1) & - & 0.008 & - & 0.012 & - & 0.983 & -\\
\textsc{Tapir}-Trf (D2) & - & 0.004 & - & - & 0.007 & - & 0.989 \\
\textsc{Tapir}-LT & 0.031 & 0.019 & 0.948 & - & - & - & - \\
\textsc{Tapir}-LT (D1) & - & 0.010 & - & 0.016 & - & 0.973 & - \\
\textsc{Tapir}-LT (D2) & - & 0.012 & - & - & 0.019 & - & 0.972 \\
\midrule
SF-Movie \\
\cmidrule{1-1}
Ref. Model & 0.311 & 0.215 & 0.549 & 0.205 & 0.107 & 0.705 & 0.852 \\
\textsc{Tapir}-Trf & 0.079 & 0.049 & 0.895 & - & - & - & - \\
\textsc{Tapir}-Trf (D1) & - & 0.028 & - & 0.043 & - & 0.930 & -\\
\textsc{Tapir}-Trf (D2) & - & 0.021 & - & - & 0.033 & - & 0.944 \\
\textsc{Tapir}-LT & 0.084 & 0.053 & 0.889 & - & - & - & - \\
\textsc{Tapir}-LT (D1) & - & 0.041 & - & 0.069 & - & 0.906 & - \\
\textsc{Tapir}-LT (D2) & - & 0.033 & - & - & 0.053 & - & 0.922\\
\midrule
NER-CoNLL \\
\cmidrule{1-1}
Ref. Model & 0.080 & 0.057 & 0.873 & 0.038 & 0.023 & 0.928 & 0.949 \\
\textsc{Tapir}-Trf & 0.023 & 0.018 & 0.931 & - & - & - & - \\
\textsc{Tapir}-Trf (D1) & - & 0.015 & - & 0.017 & - & 0.950 & - \\
\textsc{Tapir}-Trf (D2) & - & 0.011 & - & - & 0.016 & - & 0.954 \\
\textsc{Tapir}-LT & 0.026 & 0.019 & 0.933 & - & - & - & -  \\
\textsc{Tapir}-LT (D1) & - & 0.012 & - & 0.016 & - & 0.959 & -  \\
\textsc{Tapir}-LT (D2) & - & 0.014 & - & - & 0.017 & - & 0.947 \\
\midrule
Chunk-CoNLL \\
\cmidrule{1-1}
Ref. Model & 0.063 & 0.026 & 0.885 & 0.035 & 0.027 & 0.921 & 0.934 \\
\textsc{Tapir}-Trf & 0.036 & 0.016 & 0.905 & - & - & - & - \\
\textsc{Tapir}-Trf (D1) & - & 0.014 & - & 0.028 & - & 0.920 & -\\
\textsc{Tapir}-Trf (D2) & - & 0.013 & - & - & 0.026 & - & 0.926 \\
\textsc{Tapir}-LT & 0.062 & 0.026 & 0.864 & - & - & - & -  \\
\textsc{Tapir}-LT (D1) & - & 0.019 & - & 0.036 & - & 0.905 & - \\
\textsc{Tapir}-LT (D2) & - & 0.019 & - & - & 0.041 & - & 0.901 \\
\midrule
PoS-CoNLL \\
\cmidrule{1-1}
Ref. Model & 0.122 & 0.072 & 0.788 & 0.066 & 0.050 & 0.859 & 0.885 \\
\textsc{Tapir}-Trf & 0.042 & 0.032 & 0.863 & - & - & - & - \\
\textsc{Tapir}-Trf (D1) & - & 0.025 & - & 0.033 & - & 0.890 & - \\
\textsc{Tapir}-Trf (D2) & - & 0.027 & - & - & 0.037 & - & 0.884 \\
\textsc{Tapir}-LT & 0.037 & 0.028 & 0.871 & - & - & - & -  \\
\textsc{Tapir}-LT (D1) & - & 0.023 & - & 0.031 & - & 0.898 & -  \\
\textsc{Tapir}-LT (D2) & - & 0.026 & - & - & 0.034 & - & 0.890 \\
\midrule
PoS-UD-EWT \\
\cmidrule{1-1}
Ref. Model & 0.137 & 0.076 & 0.796 & 0.049 & 0.028 & 0.907 & 0.936 \\
\textsc{Tapir}-Trf & 0.053 & 0.033 & 0.855 & - & - & - & - \\
\textsc{Tapir}-Trf (D1) & - & 0.023 & - & 0.033 & - & 0.907 & - \\
\textsc{Tapir}-Trf (D2) &  - & 0.024 & - & - & 0.034 & - & 0.905 \\
\textsc{Tapir}-LT & 0.073 & 0.046 & 0.818 & - & - & - & - \\
\textsc{Tapir}-LT (D1) & - & 0.030 & - & 0.041 & - & 0.888 & -  \\
\textsc{Tapir}-LT (D2) & - & 0.024 & - & - & 0.036 & - & 0.895 \\
\bottomrule
\end{tabular}
\end{center}
\caption{Mean of Edit Overhead, Correction Time Score and Relative Correctness. $\Delta t$ denotes delay for $t$ steps.}
\label{table:appendixincremental}
}
\end{table*}

\begin{table*}[ht] 
{\footnotesize
\captionsetup{singlelinecheck = false, justification=justified}
\setlength\tabcolsep{6pt}
\centering
\begin{tabular}{l c c c c c c c c c c c }
\toprule
& & & \multicolumn{4}{c}{Action} & & \multicolumn{4}{c}{Prefix} \\\cmidrule{4-7} \cmidrule{9-12}
Tasks & \bigcell{c}{Overall \\ \texttt{WRITE}} & \bigcell{c}{Overall \\ \texttt{REVISE}} & $\frac{R \cap C}{R}$ & $\frac{R \cap I}{R}$ & $\frac{W \cap I}{W}$ & $\frac{W \cap C}{W}$ & & $\frac{R \cap C}{C}$ & $\frac{W \cap C}{C}$ & $\frac{W \cap I}{I}$ & $\frac{R \cap I}{I}$\\
\midrule
SF-SNIPS & 0.786 & 0.214 & 0.606 & 0.394 & 0.117 & 0.883 & & 0.158 & 0.842 & 0.521 & 0.479 \\
SF-ARW    & 0.802 & 0.198 & 0.741 & 0.259 & 0.040 & 0.960 & & 0.160 & 0.840 & 0.385 & 0.615  \\
SF-Movie    & 0.729	& 0.271	& 0.655	& 0.345	& 0.075	& 0.925	& & 0.209 & 0.791 & 0.370 & 0.630 \\
NER-CoNLL    & 0.896 & 0.104 & 0.654 & 0.346 & 0.101 & 0.899 & & 0.078 & 0.922 & 0.716 & 0.284  \\
Chunk-CoNLL    & 0.773 & 0.227 & 0.650 & 0.350 & 0.168 & 0.832 & & 0.187 & 0.813 & 0.619 & 0.381  \\
PoS-CoNLL  & 0.866 & 0.134 & 0.523 & 0.477 & 0.234 & 0.766 & & 0.096 & 0.904 & 0.761 & 0.239   \\
PoS-UD-EWT  & 0.798	 & 0.202 & 0.578 & 0.422 & 0.257 & 0.743 & & 	0.165 & 0.835 & 0.706 & 0.294  \\

\bottomrule
\end{tabular}
\caption{Overall distribution of actions and prefixes on test sets using \textsc{Tapir}-Trf. $W$ represents \texttt{WRITE} and $R$ represents \texttt{REVISE}. $C$ and $I$ denote correct and incorrect output prefixes, respectively.}
\label{table:revisionreliability}
}
\end{table*}

\clearpage

\begin{table*}[ht] 
{\scriptsize
\captionsetup{singlelinecheck = false, justification=justified}
\setlength\tabcolsep{4pt}
\begin{subtable}[h]{\textwidth}
\centering
\begin{tabular}{c c c c c c c c c c c c c c c c c c c c c}
\toprule
& & \multicolumn{4}{c}{SF-SNIPS} & & \multicolumn{4}{c}{SF-ARW} & &  \multicolumn{4}{c}{SF-Movie} && \multicolumn{4}{c}{NER-CoNLL} \\ \cmidrule{3-6} \cmidrule{8-11} \cmidrule{13-16} \cmidrule{18-21}
$\tau$ & & EO & CT & RC & F1 & & EO & CT & RC & F1 & & EO & CT & RC & F1 && EO & CT & RC & F1\\
\midrule
0.0 & & 0.181 & 0.100 & 0.750 & 0.911 & & 0.079 & 0.041 &  0.906 & 0.956 & & 0.311 & 0.215 & 0.549 & 0.840 && 0.080 & 0.057 & 0.873 & 0.783 \\
0.1   & & 0.102 & 0.062 & 0.863 & 0.905 & & 0.043 & 0.022 &  0.944 & 0.950 & & 0.109 & 0.066 & 0.873 & 0.838 && 0.040 & 0.035 & 0.903 & 0.761 \\
0.2   & & 0.092	& 0.059	& 0.868	& 0.903	& & 0.039 & 0.021 &  0.945 & 0.946 & & 0.097 & 0.059 & 0.881 & 0.835 && 0.032 & 0.026 & 0.918 & 0.747 \\
0.3   & & 0.086 & 0.058 & 0.870 & 0.895 & & 0.037 & 0.020 &  0.947 & 0.941 & & 0.090 & 0.055 & 0.886 & 0.832  && 0.029 & 0.022 & 0.925 & 0.742 \\
0.4   & & 0.079 & 0.056 & 0.875 & 0.888 & & 0.034 & 0.020 &  0.948 & 0.936 & & 0.083 & 0.051 & 0.891 & 0.829 && 0.025 & 0.020 & 0.929 & 0.742 \\
0.5  & & 0.074 & 0.055 & 0.876 & 0.886 & & 0.032 & 0.019 &  0.950 & 0.934 & & 0.079 & 0.049 & 0.895 & 0.828  && 0.023 & 0.018 & 0.931 & 0.741 \\
0.6  & & 0.065	 & 0.053 & 0.878 & 0.887 & & 0.030 & 0.018  & 	0.952 & 0.930 & & 0.074 & 0.046 & 0.900 & 0.826  && 0.022 & 0.016 & 0.935 & 0.742 \\
0.7  & & 0.055	 & 0.046 & 0.895 & 0.875 & & 0.027 & 0.017  & 	0.953 & 0.926 & & 0.068 & 0.043 & 0.904 & 0.822  && 0.021 & 0.015 & 0.938 & 0.743 \\
0.8  & & 0.045	 & 0.041 & 0.906 & 0.864 & & 0.024 & 0.015  & 	0.956 & 0.918 & & 0.060 & 0.038 & 0.914 & 0.817  && 0.019 & 0.014 & 0.943 & 0.742 \\
0.9  & & 0.028	 & 0.028 & 0.934 & 0.849 & & 0.019 & 0.013  & 	0.962 & 0.908 & & 0.049 & 0.034 & 0.925 & 0.804  && 0.016 & 0.012 & 0.949 & 0.739 \\
1.0  & & 0.000	 & 0.000 & 1.000 & 0.814 & & 0.000 & 0.000  & 	1.000 & 0.894 & & 0.000 & 0.000 & 1.000 & 0.768  && 0.000 & 0.000 & 1.000 & 0.700 \\

\bottomrule
\end{tabular}
\vspace*{.5cm}
\end{subtable}
\hfill
\begin{subtable}[h]{\textwidth}
\centering
\begin{tabular}{c c c c c c c c c c c c c c c c}
\toprule
& & \multicolumn{4}{c}{Chunk-CoNLL} & & \multicolumn{4}{c}{PoS-CoNLL} & &  \multicolumn{4}{c}{PoS-UD-EWT} \\ \cmidrule{3-6} \cmidrule{8-11} \cmidrule{13-16}
$\tau$ & & EO & CT & RC & F1 & & EO & CT & RC & Acc & & EO & CT & RC & Acc\\
\midrule
0.0 & & 0.063 & 0.026 & 0.885 & 0.883 & & 0.122 & 0.072 &  0.788 & 0.923 & & 0.137 & 0.076 & 0.796 & 0.921 \\
0.1   & & 0.054 & 0.022 & 0.890 & 0.878 & & 0.085 & 0.057 &  0.816 & 0.916 & & 0.085 & 0.049 & 0.824 & 0.915  \\
0.2   & & 0.048	& 0.020	& 0.895	& 0.875	& & 0.069	& 0.047	&  0.827 & 0.916 & & 0.075 & 0.044 & 0.830 & 0.913 \\
0.3   & & 0.043 & 0.018 & 0.900 & 0.873 & & 0.059 & 0.043 &  0.836 & 0.915 & & 0.066 & 0.040 & 0.840 & 0.912  \\
0.4   & & 0.039 & 0.016 & 0.903 & 0.871 & & 0.050 & 0.037 &  0.850 & 0.914 & & 0.059 & 0.036 & 0.848 & 0.910  \\
0.5  & & 0.036 & 0.016 & 0.905 & 0.869 & & 0.042 & 0.032 &  0.863 & 0.913 & & 0.053 & 0.033 & 0.855 & 0.908   \\
0.6  & & 0.033	 & 0.015 & 0.908 & 0.868 & & 0.035 & 0.028  & 	0.878 & 0.912 & & 0.049 & 0.031 & 0.861 & 0.908  \\
0.7  & & 0.031	 & 0.014 & 0.907 & 0.867 & & 0.029 & 0.024  & 	0.893 & 0.912 & & 0.045 & 0.029 & 0.868 & 0.907  \\
0.8  & & 0.028	 & 0.013 & 0.911 & 0.868 & & 0.023 & 0.020  & 	0.910 & 0.911 & & 0.040 & 0.026 & 0.877 & 0.906  \\
0.9  & & 0.025	 & 0.012 & 0.914 & 0.866 & & 0.016 & 0.015  & 	0.934 & 0.909 & & 0.035 & 0.025 & 0.885 & 0.904  \\
1.0  & & 0.000	 & 0.000 & 1.000 & 0.833 & & 0.000 & 0.000  & 	1.000 & 0.904 & & 0.000 & 0.000 & 1.000 & 0.880  \\

\bottomrule
\end{tabular}
\end{subtable}
\caption{Incremental and non-incremental performance of \textsc{Tapir}-Trf with varying threshold $\tau$ for reproducibility purpose.}
\label{table:rawnumbers_threshold}
}
\end{table*}

\end{document}